%% file: ms.tex
\documentclass[lettersize,journal]{IEEEtran}

\usepackage{amsmath,amsfonts}
\usepackage{algorithmic}
\usepackage{algorithm}
\usepackage{array}
\usepackage[caption=false,font=normalsize,labelfont=sf,textfont=sf]{subfig}
\usepackage{textcomp}
\usepackage{stfloats}
\usepackage{siunitx}
\usepackage{url}
\usepackage{verbatim}
\usepackage{graphicx}
\usepackage{cite}
\usepackage{booktabs}
\usepackage{tabularx}
\usepackage{multirow}
\usepackage{xcolor}
\usepackage{soul}
\usepackage{amssymb}
\usepackage{hyperref}
\hypersetup{
pdftitle={HyperE2VID: Improving Event-Based Video Reconstruction via Hypernetworks},
pdfsubject={Computer Vision, Event Vision},
pdfauthor={Burak Ercan, Onur Eker, Canberk Saglam, Aykut Erdem, Erkut Erdem},
pdfkeywords={Event cameras, Dynamic Vision Sensor, Event-based Vision, Video Reconstruction, Dynamic Neural Networks, Hypernetworks, Dynamic Convolutions}
}
\input{my-shortcuts.tex}

\begin{document}

\title{HyperE2VID: Improving Event-Based Video Reconstruction via Hypernetworks}

\author{Burak Ercan\thanks{BE is with the Department of Computer Engineering, Hacettepe University, TR-06800 Ankara, Turkey, also with HAVELSAN Inc., TR-06510, Ankara, Turkey}, Onur Eker\thanks{OE is with the Department of Computer Engineering, Hacettepe University, TR-06800 Ankara, Turkey, also with HAVELSAN Inc., TR-06510, Ankara, Turkey}, Canberk Saglam\thanks{CS is with the Department of Computer Engineering, Hacettepe University, TR-06800 Ankara, Turkey, also with ROKETSAN Inc., TR-06780, Ankara, Turkey}, Aykut Erdem\thanks{AE is with the Department of Computer Engineering, Koc University, TR-34450 Istanbul, Turkey, also with the Koc University Is Bank AI Center, TR-34450, Istanbul, Turkey.},~\IEEEmembership{Senior Member,~IEEE}, Erkut Erdem\thanks{EE is with the Department of Computer Engineering, Hacettepe University, TR-06800 Ankara, Turkey.}~\IEEEmembership{Senior Member,~IEEE}
\thanks{This paper has supplementary downloadable material as a single PDF file, available at http://ieeexplore.ieee.org., provided by the author. Contact burakercan@hacettepe.edu.tr for further questions about this work.}
}



\maketitle

\input{chapters/00_abstract.tex}
\input{chapters/01_intro.tex}
\input{chapters/02_related.tex}
\input{chapters/03_approach.tex}

\input{chapters/04_experiments.tex}

\input{chapters/05_conclusion.tex}

\section{Acknowledgments}
This work was supported in part by KUIS AI Research Award, TUBITAK-1001 Program Award No. 121E454,  Hacettepe University BAP Coordination Unit with grant no. FHD-2023-20611, and BAGEP 2021 Award of the Science Academy to A. Erdem. We thank all the reviewers for their valuable comments.

\bibliographystyle{IEEEtran}
\bibliography{ms}

\newpage

\vfill
\end{document}


\title{Supplementary Material for \\ ``HyperE2VID: Improving Event-Based Video Reconstruction via Hypernetworks"}

\author{Burak Ercan, Onur Eker, Canberk Saglam, Aykut Erdem,~\IEEEmembership{Senior Member,~IEEE}, Erkut Erdem,~\IEEEmembership{Senior Member,~IEEE},}

\maketitle

In this supplementary document, we provide additional material to complement the main paper. First, we provide comprehensive information about the datasets employed in our evaluations and their specific roles in our analysis. Following that, we offer the implementation details of the evaluation metrics to aid in reproducibility. Next, we delve into the detailed results of our ablation studies and further analyses. Those involve assessing various design elements of HyperE2VID and analyzing reconstruction performance in diverse scenarios, such as employing different event grouping strategies, generating high frame rate videos, and reconstructing video frames during motionless periods. Finally, we present a post-processing framework that can optionally be applied to reconstructions of any event-based video reconstruction method, eliminating or minimizing various types of artifacts encountered in textureless regions.

\section{Testing Datasets}

To comprehensively evaluate our method, we utilize sequences from five real-world datasets, namely the Event Camera Dataset (ECD)~\cite{mueggler2017event}, the Multi Vehicle Stereo Event Camera (MVSEC) dataset~\cite{zhu2018multivehicle}, the High-Quality Frames (HQF) dataset~\cite{stoffregen2020reducing}, the UZH-FPV Drone Racing dataset~\cite{delmerico2019we}, and the Color Event Camera Dataset (CED)~\cite{Scheerlinck19cvprw}. We provide the details of these datasets and describe their significance in our analysis below.

\noindent \textbf{Event Camera Dataset (ECD).} This dataset is captured by a DAVIS240C sensor \cite{brandli2014240} where events and frames are generated from the same pixel array of $240 \times 180$ resolution. Following the common practice established by Rebecq \textit{et al.} \cite{rebecq2019high}, we use seven short sequences from this dataset, where the camera moves with 6-DOF and with increasing speed in six of them. These sequences mostly contain simple office environments with static objects. The ground truth intensity frames are available at an average rate of \SI{22}{\hertz}, and we exclude scores from the initial few seconds of each sequence to align with prior work. Additionally, we exclude the parts of the sequences that contain motion blur due to fast camera motion while evaluating with full-reference metrics. In total, we use 1853 ground truth frames for full-reference metrics, and the specific start and end times of evaluation intervals are provided in \cite{stoffregen2020reducing}. We report these evaluation scores under the name \textbf{ECD} in our quantitative results tables. To assess the quality of the reconstructions under fast camera motion, we conduct a separate evaluation using the latter parts of the ECD sequences and a no-reference metric. It comprises a total of 4453 reconstructed frames, and its purpose is to examine the reconstruction quality when the camera moves rapidly. We report the scores from this evaluation under the name \textbf{FAST} in the quantitative results tables.

\vspace{0.125cm}
\noindent \textbf{Multi Vehicle Stereo Event Camera (MVSEC) dataset.} This dataset has longer sequences of indoor and outdoor environments captured by a pair of DAVIS 346B cameras. These cameras generate events and frames from the same pixel array, which has a resolution of $346 \times 260$. We use the data from the left DAVIS camera in our experiments. Following \cite{stoffregen2020reducing}, we also use specific time intervals of 6 sequences. Four of them are indoor sequences that are taken from a flying hexacopter, while the two outdoor sequences are taken from a vehicle driving in daylight. The average rate of ground truth intensity frames is around \SI{30}{\hertz} for indoor sequences and \SI{45}{\hertz} for outdoor sequences. The specific start and end times of evaluation intervals are given at \cite{stoffregen2020reducing}. The total number of ground truth frames used for evaluation is 11312. We report these scores under the name \textbf{MVSEC} in quantitative results tables. 
To assess the quality of our reconstruction method in low-light conditions, we evaluate it on the three night driving sequences from the MVSEC dataset using a no-reference metric. This evaluation comprises a total of 9415 reconstructed frames, with an average rate of \SI{10}{\hertz}. The scores from this evaluation are reported under the name \textbf{NIGHT} in the quantitative results tables.

\vspace{0.125cm}
\noindent\textbf{High-Quality Frames (HQF) dataset.} HQF dataset has 14 indoor and outdoor image sequences that cover a diverse range of motions. The data is captured by two different DAVIS240C cameras with varying noise and contrast threshold characteristics. Both cameras generate events and frames from the same $240 \times 180$ pixel array. The camera parameters and scenes are carefully selected to ensure that the ground truth frames are well-exposed and minimally motion-blurred. The dataset provides ground truth intensity frames with an average rate of \SI{22.5}{\hertz}, and following \cite{stoffregen2020reducing}, we use the entire sequences for evaluation. In total, we use 15498 ground truth frames for this evaluation. We report these evaluation scores under the name \textbf{HQF} in our quantitative results tables. We also consider a specifically curated subset of the HQF dataset to assess the quality of reconstructions under slow motion, which also poses challenges for event-based video reconstruction due to the reduced event rate. This subset, which we denote as \textbf{SLOW} and specifically utilize to analyze the effect of context information, includes all 2333 ground truth frames from two sequences named \texttt{desk\_slow} and \texttt{slow\_hand}, which were collected with the explicit aim of incorporating slow-motion scenarios.

\vspace{0.125cm}
\noindent\textbf{UZH-FPV Drone Racing (FPVDR) dataset}. This dataset is captured by a miniDAVIS346 (mDAVIS) camera mounted on a quadrotor flown by an expert drone racing pilot with fast and aggressive movements. The dataset consists of 26 indoor and outdoor flight sequences, with a total flight distance of more than \SI{10}{\kilo\meter}. The events and frames are generated from the same $346 \times 260$ pixel array of mDAVIS, which is positioned either forward facing or 45-degree downward facing for each flight. We use this dataset to assess the quality of the reconstructions under fast camera motion, using a no-reference metric. We exclude the first few seconds of each sequence to start quantitative evaluation after the drone takes off. We use event groups that span \SI{40}{\milli\second} and evaluate the total 31067 reconstructed frames.

\vspace{0.125cm}
\noindent\textbf{Color Event Camera Dataset (CED).} This dataset consists of frames and events collected with a Color-DAVIS346~\cite{taverni2018front} camera, at $346 \times 260$ resolution. We use a few sequences, including simple objects with vibrant colors and scenes with challenging lighting conditions, to present visual results of color reconstructions.

\section{Evaluation Metrics}

\noindent \textbf{MSE.} Mean squared error is a standard metric without parameters. The only thing that can affect the result of MSE while comparing two images is the range of pixel values that images have. We use floating point pixel values in the range {\lbrack 0,1\rbrack} to calculate MSE. Lower MSE scores are better.
    
\vspace{0.125cm}
\noindent \textbf{SSIM.} For structural similarity, we use the implementation from the scikit-image library~\cite{van2014scikit}, v0.19.3. We adjusted the parameters to use the Gaussian weighting scheme described in the original paper~\cite{wang2004image}. Similar to MSE, we input images with floating point pixel values in the range {\lbrack 0,1\rbrack} to SSIM calculation. Higher SSIM scores are better.

\vspace{0.125cm}
\noindent \textbf{LPIPS.} For LPIPS~\cite{zhang2018unreasonable} we use v0.1.4 of the official implementation\footnote{\url{https://github.com/richzhang/PerceptualSimilarity}} with pre-trained AlexNet \cite{krizhevsky2017imagenet} network, which requires normalizing the images so that their pixel values are in the range {\lbrack -1,1\rbrack}. Lower LPIPS scores are better.

\vspace{0.125cm}
\noindent \textbf{BRISQUE.} For BRISQUE~\cite{mittal2012no}, we use the implementation in  \mbox{IQA-PyTorch\footnote{\url{https://github.com/chaofengc/IQA-PyTorch}}} toolbox~\cite{pyiqa}, v0.1.5, with default settings. The implementation supports 3-channel RGB images; thus, we convert intensity images into RGB images by concatenating three copies of the grayscale image along the third dimension before calculating the scores. The pixel values are again in the range of {\lbrack 0,1\rbrack}. Lower BRISQUE scores are better.

\section{Ablation Study and Further Analysis}

In the following ablation studies, we evaluate various design elements of the HyperE2VID model to verify their impact on performance. This includes a detailed comparison against the E2VID+ network~\cite{stoffregen2020reducing}, which shares similarities with our base network and employs the same training data. Specifically, we retrain E2VID+ with the same hyperparameters as HyperE2VID to assess the influence of these parameters independently of our hypernetwork architecture. We further investigate the role of previous reconstructions by modifying the E2VID+ architecture to include them. Additionally, we compare our context-guided per-pixel dynamic convolutions with standard dynamic convolutions, confirming the superiority of our approach.

A significant part of our ablation study focuses on the use of context information. We experiment with networks using only event voxel grids as context, only previous reconstructions as context, or a combination of both, along with variations in curriculum learning and convolutional context fusion. We then analyze the effect of using different event grouping strategies. Finally, we evaluate the quality of reconstructions in two other challenging scenarios: high frame rate video generation (200 to 5000 FPS) and reconstruction during motionless periods.

\vspace{0.125cm}
\noindent\textbf{Training Settings.} We retrained E2VID+ using the same setup and hyperparameters as HyperE2VID to test if E2VID+ could benefit from our hyperparameter choices, without our hypernetwork architecture. The results, shown in the second row of Table~\ref{tab:ablation_hyper}, reveal mixed~outcomes. While the retrained E2VID+ shows improvements with respect to the original one in the ECD, FAST, NIGHT, and FPVDR datasets, it falls short in the MVSEC and HQF datasets. This inconsistency suggests that the enhancements are not solely due to optimizing the hyperparameters. A direct comparison with HyperE2VID, under identical conditions, clearly shows the superiority of our hypernetworks-based~approach.

\vspace{0.125cm}
\noindent\textbf{Previous Reconstructions.} In another experiment, we modify the E2VID+ architecture to include reconstructed intensity image from the previous timestep ($\hat{I}_{k-1}$) along with the current event tensor ($V_k$) via concatenation at the input. This is to distinguish the benefits of our architectural features from the simple use of past reconstructions. Even with the addition of curriculum learning, similar to HyperE2VID, this variant (shown in the third row of Table~\ref{tab:ablation_hyper}) underperforms compared to both the standard and retrained E2VID+. This highlights the unique effectiveness of our hypernetworks and dynamic per-pixel convolutions.

\input{tables_supp/ablation_hyper}

\noindent\textbf{Dynamic Convolutions.} We also compare our context-guided per-pixel dynamic convolutions with standard dynamic convolutions that lack these features. Training networks with dynamic convolutions~\cite{chen2020dynamic} or CondConv~\cite{yang2019condconv} instead of the proposed CGDD block leads to a significant drop in performance, as shown in the fourth and fifth rows of Table~\ref{tab:ablation_hyper}. It highlights the effectiveness of the proposed context-guided per-pixel dynamic convolutions in HyperE2VID in enhancing reconstruction quality.

\input{tables_supp/ablation_context_combined}

\vspace{0.125cm}
\noindent\textbf{Context Information.} 
 Moreover, we carry out several ablation experiments in order to evaluate the design choices regarding the context information used for guiding the dynamic filter generation process in HyperE2VID. Specifically, we investigate hypernetworks that use only event voxel grids as context, only previous reconstructions as context, or a combination of both, denoted as EVG, PR, and EVG+PR, respectively. It should be emphasized that these HyperE2VID variants specifically modify the context tensor computation within the CF block, while maintaining the event tensor at the input of the head layer and preserving the dynamic network architecture of both the DFG and CGDD blocks. For EVG+PR, we also examine the impact of using the curriculum learning strategy (CL) and convolutional context fusion (CF). When CF is not used, we concatenate the previously reconstructed images and event tensors channel-wise and downsample the resulting tensor to match the input of the dynamic convolution in the CGDD block. The results are summarized in Table~\ref{tab:ablation_context}. Here, we also give the results on the SLOW subset of the HQF dataset, containing slow motion.

\input{figures_supp/slow_fast_im_ev_qual}

Our quantitative results highlight the significance of choosing the right context based on the scene's characteristics. For instance, in slow-motion scenarios (SLOW), the network utilizing solely previous reconstructions (PR) vastly outperforms the one using only event voxel grids as context (EVG). Conversely, in scenes with fast motion (FAST, FPVDR) or low light conditions (NIGHT), PR's performance diminishes. To visually illustrate these findings, Fig.~\ref{fig:sfieq} shows two representative scenes from our test datasets. The first scene, from the FAST segment of the ECD dataset, highlights the limitations of standard camera intensity frames for fast motion, which struggle with either motion blur or underexposure, while the event data adeptly captures the dynamic edges of the scene. This effectively demonstrates the strength of event data in high-speed conditions. The second scene, from the SLOW segment of the HQF dataset, presents a slow-motion environment where intensity frames capture detailed visual information, but events are generated sparsely, capturing only significant brightness changes. Consequently, much of the visual detail in the scene is not visible in the event data.

Our findings in Table~\ref{tab:ablation_context} also reveal that leveraging both events and previous reconstructions as contextual information (EVG+PR) generally outperforms using only events (EVG) or only reconstructions (PR) as context. When using both events and reconstructions (EVG+PR), incorporating only the context fusion (CF) yields performance improvements on the MVSEC dataset. In contrast, incorporating only the curriculum learning strategy (CL) enhances performance on both the MVSEC and HQF datasets. Combining all these components results in our proposed HyperE2VID model (last row), which achieves the best scores on ECD, MVSEC, and HQF datasets, and second-best scores on FAST and NIGHT datasets. The variant using only event voxel grids as context (EVG), despite struggling on the HQF dataset and especially in its slow-motion sequences, excels in the fast-motion and night driving sequences. This is also visible in the top row of Fig.~\ref{fig:sfieq}, where the reconstruction of EVG is sharper and has minimal artifacts, even compared to the reconstruction of HyperE2VID. While HyperE2VID achieves the highest scores overall, the superior performance of the event-only model in certain scenarios suggests potential room for improvement for our context fusion block for future work.

\vspace{0.125cm}
\noindent\textbf{Event Grouping.}  
We perform additional experiments to assess the effect of using different event grouping strategies, that is, forming event voxel grids with different temporal windows and event numbers. First, we investigate the case with fixed temporal window grouping and conduct ten sets of experiments, each utilizing a different temporal window ranging from 10 ms to 100 ms. Second, we examine the case of fixed number event grouping and perform ten additional sets of experiment runs, each employing fixed number event grouping with a different event count ranging from 2K to 45K. For each experiment set, we consider the four best-performing methods (E2VID+, FireNet+, ET-Net, and HyperE2VID) and reconstruct videos with them using events from ECD, MVSEC, and HQF datasets by utilizing the selected event grouping strategy. For all these experiments, we employ a tolerance of 1 ms to match the reconstructions with ground truth frames and calculate LPIPS scores whenever there is a match. We then compute mean LPIPS scores for each method and for each experiment. The results, given in Fig.~\ref{fig:robustness_plots}, demonstrate the superiority of the proposed HyperE2VID architecture for generating high-quality reconstructions over a wide range of event grouping settings.

\vspace{0.125cm}
\noindent\textbf{High Frame Rate Video Reconstruction.} For high frame rate video reconstruction, Rebecq~\etal~\cite{rebecq2019high} suggested a method that groups a fixed number of events and runs multiple reconstructions in parallel, each with a slight temporal shift. This technique, however, necessitates the selection of an event count and a temporal shift value. This involves conducting numerous separate reconstructions to produce a set of videos, which are then merged by reordering frames and subjected to temporal filtering to mitigate flickering, ultimately yielding a video with a variable frame rate. In contrast, our approach utilizes fixed-temporal-window event grouping without the need for temporal shifts or parallel reconstructions, facilitating the generation of a high and constant frame rate video. The temporal window is straightforwardly determined based on the desired frame rate, using the formula 1/FPS, where a smaller window correlates with a higher FPS. This simplistic  method reveals that most event-based video reconstruction networks from existing literature begin to falter in visual quality when the FPS exceeds one thousand, as the event voxel grid statistics start to diverge from the conditions they were trained under. HyperE2VID, however, consistently produces high-contrast, sharp reconstructions, even at frame rates of several thousand frames per second. Fig.~\ref{fig:high_fps} presents reconstructed videos at high frame rates, ranging from 200 FPS to 5000 FPS. Owing to its dynamic network architecture, HyperE2VID adeptly adjusts to the varying event statistics, thus maintaining superior visual quality in high FPS video output.

\vspace{0.125cm}
\noindent\textbf{Reconstruction During Still Periods.} 
Another challenging case for event-based video reconstruction is the stationary sections in event sequences since the event rate drastically reduces, with only noise events being generated by the camera. Here, we qualitatively analyze the reconstruction quality of HyperE2VID and other methods during these motionless periods by presenting their reconstructions in Fig.~\ref{fig:pause}. The desired functionality for methods is to retain their most recent reconstructions during the pause segment, but most of them start to generate intensity images with degraded quality within a few seconds. On the other hand, the results presented in the last row of Fig.~\ref{fig:pause} demonstrate HyperE2VID's ability to preserve its high contrast and sharp reconstructions during the motionless segments, thanks to its dynamic network architecture, allowing it to adapt to highly varying event data.

\input{figures_supp/robustness_plot}

\input{figures_supp/high_FPS_recons}

\input{figures_supp/pause_recons}

\section{Post-Processing}

Here, we describe a post-processing procedure and present its qualitative results, which can optionally be applied to reconstructions of any event-based video reconstruction method, eliminating or minimizing various types of artifacts that might be encountered in textureless regions.

Our post-processing procedure consists of three steps: i) obtaining a filtered version of the reconstruction, ii) obtaining a soft mask to segment textureless regions from other regions, and iii) blending the original reconstruction with the filtered version of it using the soft mask. In the first step, we aim to use a simple filter that can remove artifacts in textureless regions but can also degrade image quality in other regions. We use a simple median filter with a kernel size of $3\times3$ for this, but one can use other filters according to the type of artifacts targeted. In the second step, we aim to obtain a soft mask that can segment textureless scene regions. For this, we choose to accumulate incoming events over a 2D image and decay them exponentially with time to give less weight to events further in the past, similar to a time-surface\cite{lagorce2016hots}. Since events are mostly generated from textured regions, this event image gives us a good approximation to segment textureless areas. We then apply dilation and Gaussian blur with $5\times5$ kernels to this event image, to obtain the final soft mask for that time step. We then use this mask to simply blend the original reconstruction with the filtered version of it, giving more weight to the latter for textureless regions.

\input{figures_supp/post_process}

The qualitative results of this procedure are presented in Fig.~\ref{fig:post_process}. We consider reconstructions of three models, E2VID+, ET-Net, and HyperE2VID, and show the effect of applying post-processing on them, using two scenes from the ECD and HQF datasets. Here, it can be seen that the described procedure generates visually pleasing images, by removing or minimizing most of the artifacts, such as checkerboard patterns. While it effectively removes fine-scale artifacts, the larger-scale artifacts remain, such as the ones in ET-Net's reconstruction in desk\_slow sequence, since we only employ a simple median filter with a small kernel size of $3\times3$ pixels. Although simple, the presented post-processing procedure can improve the visual results of event-based video reconstruction models in certain scenarios, indicating the potential for improvements in future work.

\bibliographystyle{IEEEtran}
\bibliography{supplement}

\newpage

\vfill

%% file: my-shortcuts.tex
\usepackage{xspace}
\usepackage{soul}

\makeatletter
\DeclareRobustCommand\onedot{\futurelet\@let@token\@onedot}
\def\@onedot{\ifx\@let@token.\else.\null\fi\xspace}

\def\ie{\emph{i.e}\onedot}

\def\etal{\emph{et al}\onedot}
\makeatother

\newcommand{\XYZ}[1]{\left\{0,\ldots,{#1}-1\right\}}
\newcommand{\XYZC}[1]{\left\{1,\ldots,{#1}\right\}}

%% file: chapters/00_abstract.tex
\begin{abstract}

Event-based cameras are becoming increasingly popular for their ability to capture high-speed motion with low latency and high dynamic range. However, generating videos from events remains challenging due to the highly sparse and varying nature of event data. To address this, in this study, we propose HyperE2VID, a dynamic neural network architecture for event-based video reconstruction. Our approach uses hypernetworks to generate per-pixel adaptive filters guided by a context fusion module that combines information from event voxel grids and previously reconstructed intensity images. We also employ a curriculum learning strategy to train the network more robustly. Our comprehensive experimental evaluations across various benchmark datasets reveal that HyperE2VID not only surpasses current state-of-the-art methods in terms of reconstruction quality but also achieves this with fewer parameters, reduced computational requirements, and accelerated inference times.

\begin{IEEEkeywords}
    Event-based vision, video reconstruction, dynamic neural networks, hypernetworks, dynamic convolutions.
\end{IEEEkeywords}

\end{abstract}

%% file: chapters/01_intro.tex
\section{Introduction} \label{sec:intro}

\IEEEPARstart{I}{n} the past decade, the field of computer vision has seen astonishing progress in many different tasks, thanks to modern deep learning methodologies and recent neural architectures. But, despite all these advances, current artificial vision systems still fall short on dealing with some \mbox{real-world} situations involving high-speed motion scenes with high dynamic range, as compared to their biological counterparts. Some of these shortcomings can be attributed to the classical frame-based acquisition and processing pipelines, since the traditional frame-based sensors have some problems such as motion blur and low dynamic range due to the underlying basic principles used for collecting light.

The recently developed \emph{event cameras} have the potential to eliminate the aforementioned issues by incorporating novel bio-inspired vision sensors which contain pixels that are asynchronous and work independently from each other~\cite{gallego2020event}. Each pixel is sensitive to local relative light intensity variations, and when this variation exceeds a threshold, they generate signals called \emph{events}, in continuous time. Therefore, the data output from these cameras is a stream of asynchronous events, where each event encodes the pixel location ($x, y$) and polarity~$p \in \{+1,-1\}$ of the intensity change, together with a precise timestamp $t$. The event stream has a highly varying rate depending on the scene details such as brightness change, motion, and texture. These working principles of event cameras bring many advantages compared to traditional frame-based cameras, such as high dynamic range, high temporal resolution, and low latency. Due to the numerous advantages it offers, event data has been increasingly incorporated into various recognition tasks, including object detection~\cite{li2022asynchronous}, semantic segmentation~\cite{jia2023event}, and fall detection~\cite{chen2022neuromorphic}. Furthermore, event data has been utilized in challenging robotic applications that require high-speed perception, such as an object-catching quadrupedal robot~\cite{forrai2023event} and an ornithopter robot capable of avoiding dynamic obstacles~\cite{rodriguez2022free}.

\input{figures/plot}

Despite its desirable properties, humans can not directly interpret event streams as we do for intensity images, and high-quality intensity images are the most natural way to understand visual data. Hence, the task of reconstructing intensity images from events has long been a cornerstone in event-based vision literature. Another benefit of reconstructing high-quality intensity images is that one can immediately apply successful frame-based computer vision methods to the reconstruction results to solve various tasks.

Recently, deep learning based methods have obtained impressive results in the task of video reconstruction from events (e.g. \cite{rebecq2019events,stoffregen2020reducing,weng2021event}). To use successful deep architectures in conjunction with event-based data, these methods typically group events in time windows and accumulate them into grid-structured representations like 3D voxel grids through which the continuous stream of events is transformed into a series of voxel grid representations. These grid-based representations can then be processed with recurrent neural networks (RNNs), where each of these voxel grids is consumed at each time step. 

Since events are generated asynchronously only when the intensity of a pixel changes, the resulting event voxel grid is a sparse tensor, incorporating information only from the changing parts of the scene. The sparsity of these voxel grids is also highly varying. This makes it hard for neural networks to adapt to new data and leads to unsatisfactory video reconstructions that contain blur, low contrast, or smearing artifacts~(\hspace{1sp}\cite{rebecq2019events,stoffregen2020reducing,cadena2021spade}). Recently, Weng \etal~\cite{weng2021event} proposed to incorporate a Transformer~\cite{vaswani2017attention} based module to an event-based video reconstruction network in order to better exploit the global context of event tensors. This complex architecture improves the quality of reconstructions, but at the expense of higher inference times and larger memory consumption.

The methods mentioned above try to process the highly varying event data with \emph{static} networks, in which the network parameters are kept fixed after training. Concurrently, there has been a line of research that investigates \emph{dynamic} network architectures that allow the network to adapt its parameters dynamically according to the input supplied at inference time. A well-known example of this approach is the notion of \emph{hypernetworks}~\cite{ha2016hypernetworks}, which are smaller networks that are used to dynamically generate weights of a larger network at inference time, conditioned on the input. This dynamic structure allows the neural networks to increase their representation power with only a minor increase in computational cost~\cite{han2021dynamic}.

\input{figures/HyperE2VID_overview.tex}

In this work, we present HyperE2VID which improves the current state-of-the-art in terms of image quality and efficiency (see Fig.~\ref{fig:plot}) by employing a dynamic neural network architecture via hypernetworks. Our proposed model utilizes a main network with a convolutional recurrent encoder-decoder architecture, similar to E2VID~\cite{rebecq2019events}. We enhance this network by employing dynamic convolutions, whose parameters are generated dynamically at inference time. These dynamically generated parameters are also spatially varying such that there exists a separate convolutional kernel for each pixel, allowing them to adapt to different spatial locations as well as each input. This spatial adaptation enables the network to learn and use different filters for static and dynamic parts of the scene where events are generated at low and high rates, respectively. We design our hypernetwork architecture in order to avoid the high computational cost of generating per-pixel adaptive filters via filter decomposition as in \cite{wang2021adaptive}. 

Fig. \ref{fig:approach_overview} presents an overview of our proposed method, HyperE2VID, for reconstructing video from events. Our approach is designed to guide the dynamic filter generation through a \emph{context} that represents the current scene being observed. To achieve this, we leverage two complementary sources of information: events and images. We incorporate a context fusion module in our hypernetwork architecture to combine information from event voxel grids and previously reconstructed intensity images. These two modalities complement each other since intensity images capture static parts of the scene better, while events excel at dynamic parts. By fusing them, we obtain a context tensor that better represents both static and dynamic parts of the scene. This tensor is then used to guide the dynamic per-pixel filter generation. We also employ a curriculum learning strategy to train the network more robustly, particularly in the early epochs of training when the reconstructed intensity images are far from optimal.

To the best of our knowledge, this is the first work that explores the use of hypernetworks and dynamic convolutions for event-based video reconstruction. The closest to our work is SPADE-E2VID~\cite{cadena2021spade} where the authors employ adaptive feature denormalization in decoder blocks of the E2VID architecture. Rather than feature denormalization, we directly generate per-pixel dynamic filters via hypernetworks for the first decoder block. Specifically, our contributions can be summarized as follows:

\begin{itemize}
\item We propose the first dynamic network architecture for the task of video reconstruction from events\footnote{Code is available at \url{https://ercanburak.github.io/HyperE2VID.html}}, where we extend existing static architectures with hypernetworks, dynamic convolutional layers, and a context fusion block.
\item We show via experiments that this dynamic architecture can generate higher-quality videos than previous state-of-the-art, while also reducing memory consumption and inference time.
\end{itemize}

%% file: figures/plot.tex
\begin{figure}[t]
    \centering
    \begin{tabular}{c}
    \includegraphics[width=0.9\columnwidth]{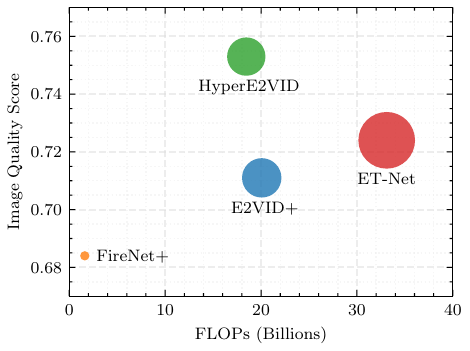}
    \end{tabular}
    \caption{\textbf{Comparison of our HyperE2VID method with state-of-the-art event-based video reconstruction methods based on image quality and computational complexity.} Image quality scores are calculated by normalizing and averaging each of the quantitative scores reported in Table~\ref{tab:quan_res}, where normalization maps the best and worst possible score for each metric to $1.0$ and $0.0$. Number of floating point operations (FLOPs) are measured as described in Section~\ref{sec:experiments_comp2}. Circle sizes indicate the number of model parameters, as detailed in Table~\ref{tab:comp}. The methods with lower image quality scores are not included for clarity of presentation.}
    \label{fig:plot}
\end{figure}

%% file: figures/HyperE2VID_overview.tex
\begin{figure}[t]
    \centering
    \begin{tabular}{c}
    \includegraphics[width=0.95\columnwidth]{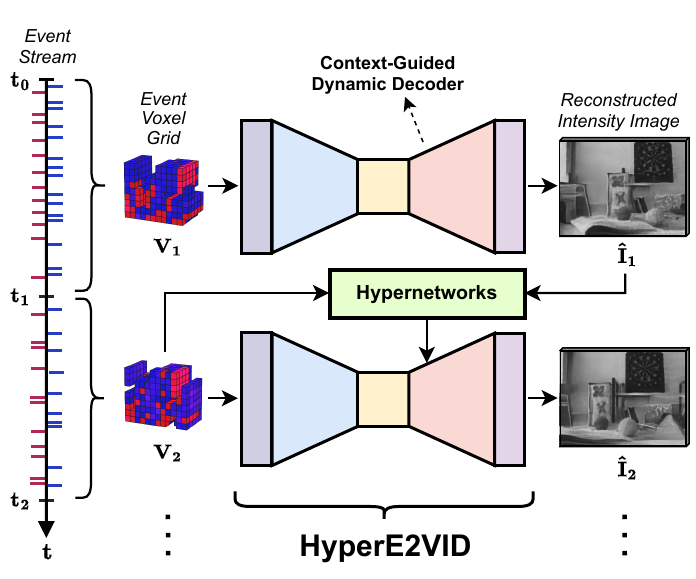}
    \end{tabular}
    \caption{\textbf{HyperE2VID} uses a recurrent encoder-decoder backbone, consuming an event voxel grid at each time step. It enhances this architecture by employing per-pixel, spatially-varying dynamic convolutions at the decoder, whose parameters are generated dynamically at inference time via hypernetworks.}
    \label{fig:approach_overview}
\end{figure}

%% file: chapters/02_related.tex
\section{Related Work} \label{sec:related}

\input{chapters/02A_related_event2im.tex}
\input{chapters/02B_related_dynamic.tex}
\input{chapters/02C_related_dynamic_ebv.tex}

%% file: chapters/02A_related_event2im.tex
\subsection{Event-Based Video Reconstruction} \label{sec:related_event2im}

Reconstructing intensity images from events is a popular topic in event-based vision literature, characterized by a variety of approaches with distinct assumptions and methodologies. Initial efforts in this field often depended on restrictive assumptions such as predetermined or limited camera movement, static scenes, or brightness constancy. More recent advancements, however, leverage deep learning techniques, which naturally integrate image priors into their models during the learning process.

Pioneering works, initiated by Cook et al.\cite{cook2011interacting}, typically aimed at simultaneously estimating multiple quantities like intensity images, spatial gradients, and optical flow~\cite{kim2014simultaneous,kim2016real,bardow2016simultaneous}. This multi-faceted approach benefits from the dynamic interaction between these elements, as exemplified by the event generation model of Gallego et al.\cite{gallego2015event}, which correlates optical flow, scene gradients, and event data. In methods that primarily predict scene gradients, a common subsequent step involves employing Poisson integration~\cite{agrawal2005algebraic} to derive intensity images from these gradients.

Kim \etal~\cite{kim2014simultaneous} introduced a filter-based method for estimating scene gradients and ego-motion, but it was limited to rotational camera movements. They later expanded this work in \cite{kim2016real} to accommodate free camera motion, though still confined to static scenes. Bardow \etal~\cite{bardow2016simultaneous} approached dynamic scenes by variational optimization, estimating intensity images and optical flow under the brightness constancy assumption.

Barua \etal~\cite{barua2016direct} were the first to show that motion estimation is not necessary for intensity image reconstruction, employing a patch-based dictionary learning method. Following a similar vein, Munda \etal~\cite{munda2018real} proposed an optimization-based method, minimizing an energy function with a data fidelity term based on direct event integration and a manifold regularization term. Scheerlinck \etal~\cite{Scheerlinck18accv} also used event integration but added a per-pixel temporal high-pass filter to mitigate noise. While their approach allowed for continuous time processing of events, it resulted in artifacts due to the loss of low-frequency information from static backgrounds.

The last few years have witnessed many works that utilize neural networks and deep learning methodologies for the task of intensity image reconstruction. Wang \etal~\cite{wang2019event_cvpr} represented groups of events with spatio-temporal voxel grids and fed them to a conditional GAN to output intensity images. In their seminal work, Rebecq \etal~\cite{rebecq2019events} proposed a recurrent fully convolutional network called E2VID to which they input voxel grids of events to produce an intensity image. They trained this network on a large synthetic dataset generated with ESIM~\cite{rebecq2018esim} using the perceptual loss of \cite{zhang2018unreasonable} and showed that this generalizes well to real event data at test time. As a follow-up study \cite{rebecq2019high}, the authors employed temporal consistency loss \cite{lai2018learning} to minimize temporal artifacts. 

After E2VID, many works attempted to enhance it from various perspectives. Scheerlinck \etal~\cite{Scheerlinck20wacv} replaced E2VID architecture with a lightweight recurrent network called FireNet, which has much less memory consumption and faster inference. However, the reconstructions of FireNet were not as good, particularly in scenarios with fast motion. Stoffregen \etal~\cite{stoffregen2020reducing} improved the results of E2VID and FireNet by matching statistics of synthetic training data to that of real-world test data, resulting in E2VID+ and FireNet+. Cadena \etal~\cite{cadena2021spade} employed spatially-adaptive denormalization (SPADE)~\cite{park2019semantic} layers in E2VID architecture, improving the quality of reconstructed videos, especially for early frames, but with an increased computational cost. Similarly, Weng \etal~\cite{weng2021event} incorporated a Transformer~\cite{vaswani2017attention} based module to the CNN-based encoder-decoder architecture of E2VID, improving the reconstruction quality at the expense of increased computational complexity.

In contrast to these, a few recent works followed somewhat different approaches, mainly targeting aspects other than the quality of reconstructions. As an example, Paredes-Vallés and de Croon~\cite{paredes2021back} turned back to the idea of simultaneously estimating optical flow and intensity images via photometric constancy assumption, and suggested a method based on self-supervised learning, eliminating the need for synthetic training data with ground truth frames. Zhu \etal~\cite{zhu2022event} used a deep spiking neural network (SNN) architecture, targeting computationally efficient neuromorphic hardware. Zhang \etal~\cite{zhang2022formulating} formulated the event-based image reconstruction task as a linear inverse problem based on optical flow, and suggested a method without training deep neural networks. Although these methods brought improvements in aspects like required training data, computational efficiency, or explainability, the visual quality of their reconstructions was not as strong.

There are also works that target a slightly different task. As an example, Zhang \etal~\cite{zhang2020learning} argued that the reconstruction performance of E2VID deteriorates when operated with low-light event data, and proposed a novel unsupervised domain adaptation network to generate intensity images as if captured in daylight, from event data of low-light scenes. Mostafavi \etal~\cite{mostafavi2020learning} presented a network to generate super-resolved intensity images from events. Similarly, Wang \etal~\cite{wang2020eventsr} introduced a network that can also perform image restoration and super-resolution. 

%% file: chapters/02B_related_dynamic.tex
\subsection{Dynamic Networks} \label{sec:related_dynamic}

Dynamic network is a generic term used to define a network that can adapt its parameters or computational graph dynamically according to its inputs at inference time \cite{han2021dynamic}. This dynamic adaptation can be accomplished in many different ways. For example, one can use a hypernetwork~\cite{ha2016hypernetworks}, which is a smaller network that is used to dynamically generate weights of a larger network conditioned on the input. For convolutional networks, dynamic filter generation can be position specific as well, such that a different filter is generated for each spatial location and the filtering operation is not translation invariant anymore~\cite{jia2016dynamic}. Position-specific dynamic filters can be pixel-wise, with a separate kernel for each spatial position, or patch-wise to reduce computational requirements. For example, Nirkin \etal~proposed HyperSeg, a semantic segmentation network~\cite{nirkin2021hyperseg} where the encoder generates parameters for dynamic patch-wise convolutional layers in the decoder. In \cite{shaham2021spatially}, Shaham \etal~proposed a Spatially-Adaptive Pixel-wise Network (ASAP-Net), where a lightweight convolutional network acts as a hypernetwork. This hypernetwork works on a lower-resolution input and produces parameters of spatially varying pixel-wise MLPs that process each pixel of the higher-resolution input independently.

It is also possible to dynamically adjust network parameters rather than directly generating them, for example by applying soft attention over multiple convolutional kernels. Both Yang \etal~\cite{yang2019condconv} and Chen \etal~\cite{chen2020dynamic} proposed to calculate a sample-specific convolutional kernel as a linear combination of many convolutional kernels, where combination coefficients are generated dynamically for each sample. Su \etal~\cite{su2019pixel} introduced Pixel-Adaptive Convolution (PAC), where they modify the spatially invariant convolutional kernel by multiplying it with a spatially varying adapting kernel that depends on the input. Chen \etal~\cite{chen2021dynamic} proposed to spatially divide the input feature into regions and process each region with a separate filter. Wang \etal~\cite{wang2021adaptive} proposed Adaptive Convolutions with Dynamic Atoms (ACDA), where they generate sample-specific convolutional filters by multiplying pixel-wise dynamic filter atoms with learned static coefficients. They also decomposed the dynamic atoms to reduce the computational requirements of calculating pixel-wise dynamic filters. 

Another approach to dynamic filters is to adapt the shape of the convolutional kernel rather than its parameters. Deformable convolution~\cite{dai2017deformable} deforms the geometric structure of the convolutional filter to allow sampling from irregular points. This is achieved by augmenting each sampling location in the filter with dynamic offsets generated by another learned convolutional kernel.

%% file: chapters/02C_related_dynamic_ebv.tex
\subsection{Dynamic Networks for Event-Based Vision} \label{sec:related_dynamic_ebv}

Recently, the concept of dynamic networks have started to be used in event-based vision literature as well. In \cite{zou2021learning}, \cite{han2021evintsr} and \cite{messikommer2022multi}, deformable convolution based feature alignment modules are used for event-based image reconstruction, super-resolution, and HDR imaging, respectively. Vitoria \etal~\cite{vitoria2023event} used modulated deformable convolutions for the task of event-based image deblurring, where event features encode the motion in the scene, in the form of kernel offsets and modulation masks. Xie \etal~\cite{Xie2022VMVGCNVM} employed dynamically updated graph CNN to extract discriminative spatio-temporal features for event stream classification.

While the aforementioned methods focus on dynamically changing the computational graphs of networks, there are also works that directly generate network parameters in a dynamic manner. For instance, in the task of event-based video super-resolution, Jing et al.\cite{jing2021turning} employed a network that takes event representations as inputs and generates parameters for dynamic convolutional layers. In contrast, we employ a context fusion mechanism and generate dynamic parameters guided by both event and image information, motivated by the complementary nature of these two domains. Xiao et al.~\cite{Xiao2022EVA2EV} used dynamic convolutional filters similar to our method but for event-based video frame interpolation. However, they applied each convolutional kernel of shape $1 \times k \times k$ to a specific feature channel to reduce computational demand, which prevents effective modeling of inter-channel dependencies. On the other hand, we consider usual 2D convolutions to let the network model these dependencies, while avoiding high computational costs by using two filter decomposition steps. Furthermore, we utilize previously reconstructed intensity images for context fusion and employ a curriculum learning strategy for robust training, as will be detailed later.

%% file: chapters/03_approach.tex
\section{The Approach} \label{sec:approach}

\input{chapters/03A_approach_formulation.tex}
\input{chapters/03B_approach_event_repr.tex}
\input{chapters/03C_approach_hyper.tex}
\input{chapters/03D_approach_training_details.tex}

%% file: chapters/03A_approach_formulation.tex
\subsection{Formulation} \label{sec:approach_formulation}

Let us assume that we have an event stream~$\{e_i\}$ consisting of $N_E$ events that span a duration of $T$ seconds. Each event ~$e_i=(x_i,y_i,t_i,p_i)$ encodes the location $x_i$ and $y_i$, the timestamp $t_i$ and the polarity $p_i$ of the $i$th brightness change that is perceived by the sensor, such that $t_i \in [0, T]$, $p_i \in \{+1,-1\}$, $x_i \in \XYZ{W}$ and $y_i \in \XYZ{H} $ for all $ i \in \XYZ{N_E}$, where $W$ and $H$ are the width and the height of the sensor array, respectively. 

Given only these events, our task is to generate an image stream~$\{ \hat{I}_k \}$ of $N_I$ images from that same time period of $T$ seconds. Each image $\hat{I}_k \in [0,1]^{W \times H}$ is a 2D grayscale representation of the absolute brightness of the scene as if captured by a standard frame-based camera at some time $s_k \in [0, T]$ for all $k \in \XYZC{N_I}$. It is important to note that we constrain our method such that each generated image only depends on past events, \ie~only $\{ e_i~|~t_i \le s_k \}$ is used to generate an image $\hat{I}_k$. This allows our method to be used in scenarios where future events are not observed yet, such as reconstructing intensity images from a continuous event camera stream in real-time.

%% file: chapters/03B_approach_event_repr.tex
\subsection{Event Representation} \label{sec:approach_event_repr}

Since each event conveys very little information regarding the scene, a common approach in event-based vision literature is to accumulate some number of events into a group, for example by considering a spatio-temporal neighborhood, and then process this group together. We also follow this approach. Assuming that the ground truth intensity frames are available together with the incoming event stream, one can group events such that every event between consecutive frames ends up in the same group. Therefore, given the frame timestamps $s_k$ for all $k \in \XYZC{N_I}$, and letting $s_0 = 0$, the set of events in the $k$th event group can be defined as follows:
\begin{equation}
\label{eq:group_between_frames}
G_k \doteq \{e_i \mid s_{k-1} \leq t_i < s_k \}
\end{equation}

To utilize deep CNN architectures for event-based data, a common choice is to accumulate grouped events into a grid-structured representation such as a voxel grid~\cite{zihao2018unsupervised}. Let $G_k$ denote a group of events that spans a duration of $\Delta T$ seconds, $T_k$ represent the starting timestamp of that duration, and $B$ be the number of temporal bins that will be used to discretize the timestamps of continuous-time events in the group. The voxel grid $V_k \in {\rm I\!R}^{W \times H \times B}$ for that group is formed such that the timestamps of the events from the group are first normalized to the range $[0, B-1]$, and then each event contributes its polarity to the two temporally closest voxels using a linearly weighted accumulation similar to bilinear interpolation:
\begin{equation}
\label{eq:event_representation_voxel}
V_k(x,y,t) = \sum_{\substack{i}} {p_i \max(0, 1-|t - t^*_i|) {\delta}(x-x_i,y-y_i)}
\end{equation}
where $\delta$ is the Kronecker delta that selects the pixel location, and $t^*_i$ is the normalized timestamp which is calculated as:
\begin{equation}
\label{eq:event_representation_ts_norm}
t^*_i = (B-1)(t_i - T_k) / (\Delta T)
\end{equation}
where, in all our experiments, we use $B=5$.

%% file: chapters/03C_approach_hyper.tex
\subsection{HyperE2VID} \label{sec:approach_hyper}

\input{figures/HyperE2VID_detailed.tex}

After representing each event group with a voxel grid, our task is to generate an image stream from the sequence of voxel grids. We use a recurrent neural network that consumes a voxel grid~$V_k$ at each time step  $k \in \XYZC{N_I}$, and generates an image~$\hat{I}_k$ corresponding to that specific moment. Specifically, we use a U-Net~\cite{ronneberger2015u} based fully convolutional architecture with recurrent encoder blocks, decoder blocks, and skip connections between them, similar to the~E2VID model~\cite{rebecq2019events} and the subsequent works of \cite{rebecq2019high}, \cite{stoffregen2020reducing}, and \cite{cadena2021spade}. Then, we augment this main architecture with hypernetworks, dynamic convolutions, and a context fusion module. We refer to the resulting architecture as HyperE2VID. 

Fig.~\ref{fig:network_architecture} shows an overview of the proposed HyperE2VID framework. Our model consists of a main network $\mathcal{F}$ and hypernetworks that generate parameters for the dynamic part of the main network. From its input to output, the network $\mathcal{F}$ consists of one head layer, three recurrent encoder blocks, two residual blocks, one context-guided dynamic decoder (CGDD) block, two standard decoder blocks, and a prediction layer. The dynamic filter generation (DFG) block and the context fusion (CF) block act as hypernetworks that generate pixel-wise dynamic filter parameters for the dynamic part of the main network, \ie~the CGDD block. 

More formally, let $S_{k}$ be the recurrent state of the network for a time step $k$, containing states $\mathbf{S^{e_n}_k}$ of the three encoder blocks, where $ n \in \{0,1,2\} $. Given the states from the previous time step, $S_{k-1}$, and the event voxel grid from the current time step, $V_k$, the main network $\mathcal{F}$ calculates the current states $S_{k}$ and predicts the intensity image $\hat{I}_{k}$ as follows:
\begin{equation}
\label{eq:hypere2vid_main}
(\hat{I}_{k}, S_k) = \mathcal{F}(V_k, S_{k-1}, \theta_k)
\end{equation}
with $ \theta_k $ denoting the parameters of the convolutional layer at the CGDD block, which are generated dynamically at inference time by the DFG block, as below:
\begin{align}
C_k &= \text{CF}(V_k, \hat{I}_{k-1}) \label{eq:ck} \\
\theta_k &= \text{DFG}(C_k) \label{eq:dfgd}
\end{align}

To generate the parameters of the dynamic decoder, we use both the current event voxel grid $V_k$ and the previous reconstruction result $\hat{I}_{k-1}$. The CF block fuses these inputs to generate a context tensor $C_k$, which is then used by the DFG block. This approach is motivated by the complementary nature of the two domains. Events are better suited for capturing fast motion due to their high temporal resolution but cannot capture static parts of the scene. In contrast, intensity images are better at capturing static parts of the scene. By fusing $V_k$ and $\hat{I}_{k-1}$, the context tensor $C_k$ incorporates useful features that better describe the static and dynamic parts of the scene.

Skip connections carry output feature maps of the head layer and each encoder block to the inputs of the respective symmetric decoder components, \ie~before each decoder block and the prediction layer. Element-wise summation is performed for these skip connections. ReLU activations are used for each convolutional layer unless specified otherwise. We describe each component of our architecture in more detail below:

\noindent\textbf{Head layer}. The head layer consists of a convolutional layer with a kernel size of 5. The convolutional layer processes the event voxel grid with 5 temporal channels and outputs a tensor with 32 channels, while the input's spatial dimensions $H$ and $W$ are maintained.

\noindent\textbf{Encoder blocks}. Each encoder block consists of a convolutional layer followed by a ConvLSTM~\cite{shi2015convolutional}. The convolutional layer has a kernel size of 5 and stride of 2, thus, it reduces the spatial dimensions of the input feature map by half. On the other hand, it doubles the number of channels. The ConvLSTM has a kernel size of 3 and maintains the spatial and channel dimensions of its inputs and internal states.

\noindent\textbf{Residual blocks}. Each residual block in our network comprises two convolutional layers with a kernel size of 3 that preserve the input's spatial and channel dimensions. A skip connection adds the input features to the output features of the second convolution before the activation function.

\noindent\textbf{Context-Guided Dynamic Decoder (CGDD) block}. The CGDD block includes bilinear upsampling to increase the spatial dimensions, followed by a dynamic convolutional layer. The convolution contains $5 \times 5$ kernels and reduces the channel size by half. The parameters $\theta_k$ of this convolution are generated dynamically during inference time by the DFG block. 

It is important to emphasize that all dynamic parameters are generated pixel-wise in that there exists a separate convolutional kernel for each pixel. This spatial adaptation is motivated by the fact that the pixels of an event camera work independently from each other. When there is more motion in one part of the scene, events are generated at a higher rate at corresponding pixels, and the resulting voxel grid is denser in those regions. Our design enables the network to learn and use different filters for each part of the scene according to different motion patterns and event rates, making it more effective to process the event voxel grid with spatially varying densities. 

\noindent\textbf{Standard Decoder blocks}. Each standard decoder block consists of bilinear upsampling followed by a standard convolutional layer. The details are the same as the context-guided dynamic decoder, except that the parameters are learned at training time and fixed at inference time.

\noindent\textbf{Prediction layer}. The prediction layer is a standard convolutional layer with a kernel size of 1, and it outputs the final predicted intensity image with 1 channel. We do not use an activation function after this layer.

\noindent\textbf{Dynamic Filter Generation (DFG) block}. \label{subsec:dfg} A crucial component of our method is the dynamic filter generation. This block consumes a \emph{context} tensor and output parameters for the CGDD block. The context tensor $C_k$ is expected to be at the same spatial size as the input of the dynamic convolution ($ W'' \times H''$). To generate the context tensor, we use a context fusion mechanism that fuses features from the event voxel grid ($V_k$) and the previous reconstruction ($\hat{I}_{k-1}$) of the network. 

To reduce the computational cost, we use two filter decomposition steps while generating per-pixel dynamic filters. First, we decompose filters into per-pixel filter atoms generated dynamically. Second, we further decompose each filter atom as a truncated expansion with pre-fixed multi-scale Fourier-Bessel bases. Inspired by ACDA~\cite{wang2021adaptive}, our approach generates efficient per-pixel dynamic convolutions that vary spatially. However, unlike ACDA, our network architecture performs dynamic parameter generation independently through hypernetworks, which are guided by a context tensor designed to provide task-specific features for event-based video reconstruction.

\input{figures/DynamicFilterGeneration.tex}

Fig.~\ref{fig:dfg} illustrates the detailed operations of our proposed DFG block. A context tensor with dimensions $ W'' \times H'' \times C_\textit{cont}$ is fed into a 2-layer CNN, producing pixel-wise basis coefficients of size $C_\textit{coeff}$ that are used to generate per-pixel dynamic atoms via pre-fixed multi-scale Fourier-Bessel bases.
These bases are represented by a tensor of size $ s \times b \times l \times l$, where $s$ is the number of scales, $b$ is the number of Fourier-Bessel bases at each scale, and $l$ is the kernel size for which the dynamic parameters are being generated. Multiplying the multi-scale Fourier-Bessel bases with the basis coefficients generate per-pixel dynamic atoms of size $l \times l$. Number of generated atoms for each pixel is $a$, so it is possible to represent all of the generated atoms by a tensor of size $W'' \times H'' \times a \times l \times l$. Next, the compositional coefficients tensor of size $C_\textit{in} \times a \times C_\textit{out}$ is multiplied with these per-pixel dynamic atoms. These learned coefficients are fixed at inference time and shared across spatial positions. This multiplication produces a tensor of size $W' \times H' \times C_\textit{in} \times C_\textit{out} \times l \times l$, which serves as the parameters for the per-pixel dynamic convolution. Here, $C_\textit{in}$ and $C_\textit{out}$ are the number of input and output channels for the dynamic convolution, respectively.
For the DFG block, we set $a=6$, $b=6$, and $l=5$. The number of scales $s=2$, meaning that we use $3 \times 3$ and $5 \times 5$ sized Fourier-Bessel bases. Since we have $b=6$ bases at each scale, we have a total of $s \times b = 12$ Fourier-Bessel bases. The 2-layer CNN has a hidden channel size of 64. Both convolutional layers have a kernel size of 3, and they are followed by a batch normalization~\cite{ioffe2015batch} layer and a tanh activation. 
The output of the CNN has $C_\textit{coeff} = a \times b \times s = 72$ channels, to produce a separate coefficient per dynamic atom and per Fourier-Bessel basis.

\noindent\textbf{Context Fusion (CF) block}. The events are generated asynchronously only when the intensity of a pixel changes, and therefore the resulting event voxel grid is a sparse tensor, incorporating information only from the changing parts of the scene. Our HyperE2VID architecture conditions the dynamic decoder block parameters with both the current event voxel grid $V_k$ and the previous network reconstruction $\hat{I}_{k-1}$. These two domains provide complementary information; the intensity image is better suited for static parts of the scene, while the events are better for dynamic parts. We use the CF block to fuse this information, enabling the network to focus on intensity images for static parts and events for dynamic parts. Our context fusion block design concatenates $V_k$ and $\hat{I}_{k-1}$ channel-wise to form a 6-channel tensor. We downsample this tensor to match the input dimensions of the dynamic convolution at the CGDD block and then use a $3 \times 3$ convolution to produce a context tensor with 32 channels. While more complex architectures are possible, we opt for a simple design for the context fusion block.

%% file: figures/HyperE2VID_detailed.tex
\begin{figure*}[ht]
\centering
\includegraphics[width=1.0\linewidth]{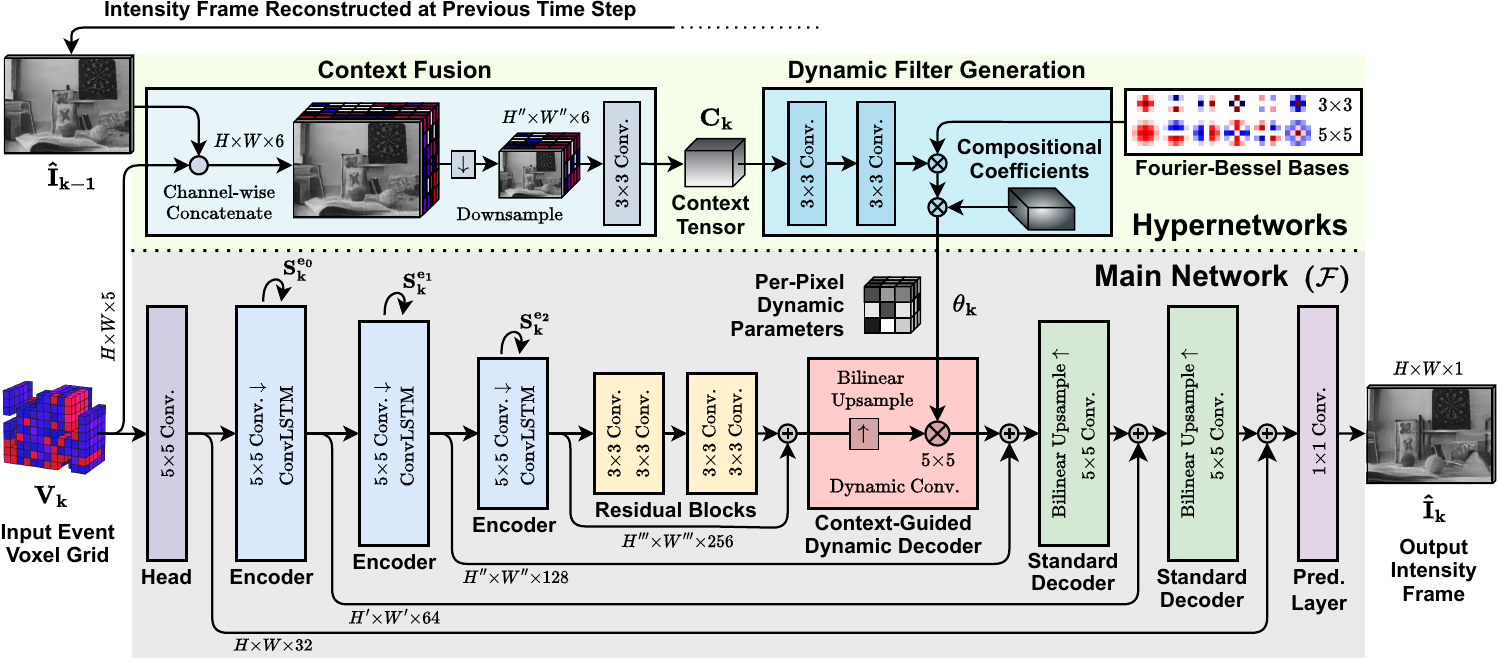}
\caption{\textbf{Overview of our proposed HyperE2VID architecture.} The main network $\mathcal{F}$ uses a U-Net like architecture to process an event voxel grid~$V_k$ and predict the intensity image $\hat{I}_{k}$ at each time step $k$. It includes downsampling encoder blocks, upsampling decoder blocks, and skip connections. The encoders incorporate ConvLSTM blocks to capture long temporal dependencies in the sparse event stream. The parameters of the context-guided dynamic decoder (CGDD) block are generated dynamically at inference time, enabling the network to adapt to highly varying event data. These parameters are generated via hypernetworks, consisting of a context fusion (CF) block and a dynamic filter generation (DFG) block. The DFG block employs two filter decomposition steps using multi-scale Fourier-Bessel Bases and learned compositional coefficients, avoiding the high computational cost of per-pixel adaptive filters. The CF block fuses event features from the current time step $k$ with reconstructed image features from the previous time step $k-1$ to generate a context tensor. This fusion scheme combines the dynamic and static parts of the scene captured by events and images, respectively, to generate a context tensor that better represents the overall scene.}
\label{fig:network_architecture}
\end{figure*}

%% file: figures/DynamicFilterGeneration.tex
\begin{figure}
\centering
\includegraphics[width=0.98\columnwidth]{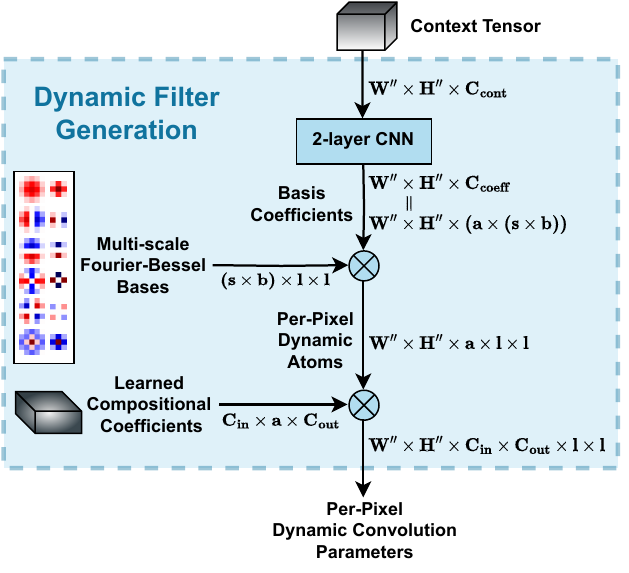}
\caption{\textbf{Dynamic Filter Generation (DFG) block}. DFG block takes a context tensor as input and generates per-pixel dynamic convolution parameters via two filter decomposition steps, making use of pre-fixed multi-scale Fourier-Bessel bases and learned compositional coefficients. More details are given in Section~\ref{subsec:dfg}.
}
\label{fig:dfg}
\end{figure}

%% file: chapters/03D_approach_training_details.tex
\subsection{Training Details} \label{sec:approach_training_details}
During training, we employ the following loss functions:

\noindent\textbf{Perceptual Reconstruction Loss.} We use the AlexNet \cite{krizhevsky2017imagenet} variant of the learned perceptual image patch similarity (LPIPS)~\cite{zhang2018unreasonable} to enforce reconstructed images to be perceptually close to ground truth intensity images. LPIPS works by passing the predicted and reference images through a deep neural network architecture that was trained for visual recognition tasks, and using the distance between deep features from multiple layers of that network as a measure of the perceptual difference between the two images.
\begin{equation}
    \mathcal{L}^{\text{LPIPS}}_k = \text{LPIPS}(\hat{I}_k, I_k)
\end{equation}

\noindent\textbf{Temporal Consistency Loss.} We use the \emph{short-term temporal loss} of \cite{lai2018learning}, as employed in \cite{rebecq2019high}, to enforce temporal consistency between the images that are reconstructed in consecutive time steps of the network. This loss works by warping the previously reconstructed image using a ground truth optical flow to align it with the current reconstruction and using a masked distance between these aligned images as a measure of temporal consistency, where the mask is calculated from the warping error between the previous and the current ground truth intensity images. More formally, the temporal consistency loss is calculated as:
\begin{equation}
    \mathcal{L}^{\text{TC}}_k = M_k \lVert \hat{I}_k - W(\hat{I}_{k-1}, F_{k \rightarrow k-1}) \lVert_1
\end{equation}
where $F_{k \rightarrow k-1}$ denotes the optical flow map between time steps $k$ and $k-1$, $W$ is the warping function, and $M_k$ represents the occlusion mask which is computed as:
\begin{equation}
    M_k = \text{exp} (- \alpha \lVert {I}_k - W({I}_{k-1}, F_{k \rightarrow k-1}) \lVert^2_2 )
\end{equation}
where we use $\alpha=50$ as in~\cite{lai2018learning,rebecq2019high}. The mask $M_k$ contains smaller terms for pixels where the warping error between consecutive ground truth images is high, and therefore the masking operation effectively discards these pixels from the temporal consistency calculation of reconstructed frames.

The final loss for a time step $k$ is the sum of the perceptual reconstruction and temporal losses:
\begin{equation}
    \mathcal{L}_k =  \mathcal{L}^{\text{LPIPS}}_k + \mathcal{L}^{\text{TC}}_k
\end{equation}

During training, we calculate the loss $\mathcal{L}_k$ at every $T_S$ time-steps in a training sequence, and the gradients of this loss with respect to the network parameters are calculated using the Truncated Back-propagation Through Time (TBPTT) algorithm \cite{williams1990efficient} with a truncation period of $T_T$ time-steps. Setting $T_S > 1$ and $T_T < k$ reduces memory requirements and speeds up the training process.

We implement our network in PyTorch \cite{paszke2019pytorch}. We train the recurrent network with sequences of length 40, with the network parameters initialized using He initialization \cite{he2015delving}. At the first time step of each sequence, the initial values of the previous reconstruction, $\hat{I}_0$, and the network states, $S_0$, are set to zero tensors. The loss calculation and the truncation periods are set as $T_S = 10$ and $T_T = 5$, respectively. We train our network for 400 epochs using a batch size of 10 and the AMSGrad~\cite{reddi2019convergence} variant of the Adam~\cite{kingma2014adam} optimizer with a learning rate of 0.001. To track our trainings and experimental analyses, we use Weights \& Biases~\cite{wandb}.

At the start of the training, the previous reconstruction $\hat{I}_{k-1}$ of the network, which is used for context fusion, is far from optimal. This makes it harder for the context fusion block to learn useful representations, especially in the earlier epochs of the training. To resolve this issue, we employ a curriculum learning~\cite{bengio2009curriculum} strategy during the training. We start the training by using the ground truth previous image $I_{k-1}$ instead of the previous reconstruction of the network $\hat{I}_{k-1}$ for context fusion. For the first 100 epochs, we gradually switch to using images that the network reconstructs at the previous time step, by weighted averaging them with ground-truth images. After the 100th epoch, we continue the training by only using previous reconstructions for context fusion. Therefore, we use a modified version of Equation~(\ref{eq:ck}) during training:
\begin{align}
\label{eq:warmup}
\beta &= \min(1,~\frac{\textit{epoch}}{100}) \\
I_{\textit{context}} &= \beta \cdot \hat{I}_{k-1} + (1 - \beta) \cdot I_{k-1} \\
C_k &= \text{CF}(V_k, I_{\textit{context}})
\end{align}

This curriculum learning strategy allows the parameters of the hypernetworks to be learned more robustly, enabling the training process to converge to a better-performing model. 

During training, we augment the images and event tensors with random crops and flips as suggested in \cite{rebecq2019high}. The size of random crops is $112 \times 112$, and the probability of vertical and horizontal flips are both 0.5. Furthermore, we employ dynamic train-time noise augmentation, pause augmentation, and hot-pixel augmentation as described in \cite{stoffregen2020reducing}.

%% file: chapters/04_experiments.tex
\section{Experimental Analysis} \label{sec:experiments}

\input{chapters/04A_experiments_trainset.tex}
\input{chapters/04B_experiments_testsets.tex}
\input{chapters/04C_experiments_metrics.tex}
\input{chapters/04D_experiments_competing.tex}
\input{chapters/04E_experiments_results.tex}

\input{chapters/04F_experiments_comp2.tex}

%% file: chapters/04A_experiments_trainset.tex
\subsection{Training Dataset} \label{sec:experiments_trainset}

We generate a synthetic training set as described in \cite{stoffregen2020reducing}, using the \texttt{Multi-Objects-2D} renderer option of ESIM~\cite{rebecq2018esim} where multiple moving objects are captured with a camera restricted to 2D motion. The dataset consists of 280 sequences, all of which are 10 secs in length. The contrast threshold values for event generation are in the range of 0.1 to 1.5. Each sequence includes generated event streams together with ground truth intensity images and optical flow maps with an average rate of \SI{51}{\hertz}. The resolutions of event and frame cameras are both $256 \times 256$. The sequences include scenes containing up to 30 foreground objects with varying speeds and trajectories, where the objects are randomly selected images from the MS-COCO dataset~\cite{lin2014microsoft}. 

%% file: chapters/04B_experiments_testsets.tex
\subsection{Testing Datasets} \label{sec:experiments_testsets}

To comprehensively evaluate our method, we utilize sequences from five real-world datasets, each selected for their unique characteristics and relevance to different aspects of event-based video reconstruction. These datasets are the Event Camera Dataset (ECD)\cite{mueggler2017event}, the Multi Vehicle Stereo Event Camera (MVSEC) dataset\cite{zhu2018multivehicle}, the High-Quality Frames (HQF) dataset~\cite{stoffregen2020reducing}, the UZH-FPV Drone Racing (FPVDR) dataset~\cite{delmerico2019we}, and the Color Event Camera Dataset (CED)~\cite{Scheerlinck19cvprw}.

The ECD dataset, with its DAVIS240C sensor-generated frames and events, is pivotal for evaluating reconstructions in environments with 6-DOF camera movement and varying speeds. Within this dataset, we introduce the FAST subset to specifically assess reconstruction quality under conditions of rapid camera motion. The MVSEC dataset offers longer sequences in both indoor and outdoor settings, captured by DAVIS 346B cameras. This dataset is integral for analyzing performance in diverse environments. Additionally, we derive the NIGHT subset from MVSEC to evaluate our method's effectiveness in low-light conditions, a challenging scenario for event-based reconstruction. The HQF dataset provides a variety of indoor and outdoor sequences with well-exposed and minimally blurred frames, crucial for benchmarking reconstruction quality in more controlled environments. The UZH-FPV Drone Racing dataset, with its fast and aggressive drone movements, is ideal for testing our method under extreme motion conditions, offering a rigorous assessment of reconstruction capabilities in dynamic scenarios. Lastly, the CED dataset's color frames and events, captured with the Color-DAVIS346 camera, allow us to demonstrate our method's color reconstruction ability, particularly in scenes with vibrant colors and challenging lighting conditions.

Detailed descriptions of these datasets and their specific usage in our analysis are given in the supplementary material.

%% file: chapters/04C_experiments_metrics.tex
\subsection{Evaluation Metrics} \label{sec:experiments_metrics}

We evaluate the methods using three full-reference evaluation metrics, mean squared error (MSE), structural similarity (SSIM)~\cite{wang2004image}, and learned perceptual image patch similarity (LPIPS)~\cite{zhang2018unreasonable} when high-quality, distortion-free ground truth frames are available. To assess image quality under challenging scenarios, such as low-light and fast motion, where ground truth frames are of low quality, we use a no-reference metric, BRISQUE~\cite{mittal2012no}. These metrics have some settings that affect their results, and thus we provide the implementation details of them in the supplementary material to facilitate reproducibility.

%% file: chapters/04D_experiments_competing.tex
\subsection{Competing Approaches} \label{sec:experiments_competing}

We compare our method against seven other methods from the literature, which are E2VID~\cite{rebecq2019high}, FireNet~\cite{Scheerlinck20wacv}, FireNet+ and E2VID+~\cite{stoffregen2020reducing}, SPADE-E2VID~\cite{cadena2021spade}, SSL-E2VID~\cite{paredes2021back}, and ET-Net~\cite{weng2021event}. E2VID+ and SSL-E2VID use the same network architecture as E2VID, but their training details are different. Similarly, FireNet+ uses the same network architecture as FireNet. We use the pre-trained models that the respective authors publicly share for each of these methods, and evaluate them using the same datasets and under the same settings. All of these methods use the same voxel grid event representation as ours (Section~\ref{sec:approach_event_repr}). We group events that have timestamps between every two consecutive ground truth frames and form the voxel grids using these. We also apply any pre-processing and post-processing steps when required by the method, such as the event tensor normalization and robust min/max normalization of E2VID. After generating reconstructions for each method, we perform quantitative analysis using the full-reference metrics, MSE, SSIM, and LPIPS, or the no-reference metric BRISQUE, depending on whether high-quality ground truth frames are available or not. We do not perform histogram equalization to reconstructions or ground truth images before calculating evaluation metrics. The quantitative results and the qualitative analysis are given in Section~\ref{sec:experiments_results}, as well as color reconstruction results for sample scenes from the CED dataset. We also compare the computational complexity of each network architecture in Section~\ref{sec:experiments_comp2}.

%% file: chapters/04E_experiments_results.tex
\subsection{Experimental Results} \label{sec:experiments_results}
\input{tables/quan_results.tex}

\input{tables/qual_eval.tex}

Table~\ref{tab:quan_res} presents the quantitative results obtained from evaluating the methods on sequences from the aforementioned datasets. We calculate the average values of each metric across all evaluated frames. The HyperE2VID method achieves state-of-the-art performance in terms of most metrics. On the ECD and MVSEC datasets, it outperforms the second-best method, ET-Net, by a large margin. On the HQF dataset, it delivers results on par with state-of-the-art approaches. In challenging scenarios involving fast camera motion (FAST and FPVDR), it obtains the best BRISQUE scores; and in night driving sequences (NIGHT), it obtains the second-best BRISQUE scores after E2VID, surpassing all the other methods. These results demonstrate the effectiveness of the proposed HyperE2VID method, which generates perceptually more pleasing and high-fidelity reconstructions.

We present qualitative results for ECD, MVSEC, HQF, and FPVDR datasets in Fig.~\ref{fig:qual_eval}. We omit reconstructions of FireNet and SPADE-E2VID due to lower quantitative scores and focus on the performances of E2VID, E2VID+, FireNet+, E2VID+, SSL-E2VID, ET-Net, and HyperE2VID. Sample scenes are shown from ECD (rows 1,2), MVSEC (rows 3,4), HQF (rows 5-7), and FPVDR (row 8) datasets, as well as fast parts of ECD (FAST, row 9) and night sequences of MVSEC (NIGHT, row 10) datasets. Each row shows reconstructions of each model (first six columns) with the reference frame given in the rightmost column.

The visual qualities of reconstructions are mostly in line with the quantitative results. Among the six methods, FireNet+ and SSL-E2VID tend to have the lowest quality, with prominent visual artifacts and blurry regions. Reconstructions of E2VID+ have fewer artifacts, especially at scenes from the HQF dataset. E2VID+ also produces nice-looking images for the outdoor scenes of the MVSEC dataset. However, its reconstructions are generally of low contrast and blurry around the edges. ET-Net has better contrast but has more artifacts at textureless regions and around the edges of objects. The reconstructions of HyperE2VID are of high contrast and sharp around the edges. Moreover, the textureless regions are mostly reconstructed with fewer artifacts. 

These qualitative results show that the reconstructions from most methods display artifacts to varying degrees. In light of this widespread issue, we present a post-processing framework in the supplementary material. This framework, which can be applied to reconstructions from any event-based video reconstruction method, aims to eliminate or significantly reduce various types of artifacts, particularly in textureless regions.

In Fig.~\ref{fig:color}, we show color reconstructions from HyperE2VID alongside those from two top-performing competitors, E2VID+ and ET-Net, using sample scenes from the CED dataset. These are compared with reference frames from the Color-DAVIS346 camera. To generate these color reconstructions, we adopt the method described in~\cite{rebecq2019high}. This involve reconstructing each color channel separately at quarter resolution, then upsampling and merging them to form a full-color image. Next, we convert this image to LAB color space and replace its luminance channel with a high-resolution grayscale reconstruction derived from all events. The color results, as shown in Fig.~\ref{fig:color}, demonstrate HyperE2VID's ability to produce color images of superior quality. These images exhibit sharp edges, minimal artifacts, and authentic colors, even in challenging lighting conditions, such as the high-dynamic-range (HDR) scene displayed in the last row.

In our supplementary material, we present comprehensive ablation studies and additional analyses of the HyperE2VID model. Key design components like context-guided per-pixel dynamic convolutions, hypernetworks, and context fusion are rigorously evaluated to affirm their impact. We specifically explore the adaptability of our method in varied scenarios including slow motion, fast motion, and low-light conditions, highlighting the critical role of contextually relevant information in these environments. Our investigation also includes the impact of varying temporal windows and event counts in constructing event voxel grids, highlighting the versatility of the HyperE2VID architecture across diverse settings. Moreover, we demonstrate its effectiveness in two particularly demanding situations: generating high frame rate videos ranging from \SI{200}{\hertz} to \SI{5}{\kilo\hertz}, and reconstructing scenes during motionless intervals. Our findings not only validate our design choices but also offer valuable directions for future enhancements.

\input{tables/color_recons}

%% file: tables/quan_results.tex
\begin{table*}[!t]
\small
\caption{Quantitative results of existing methods and our proposed method on sequences from ECD, MVSEC, HQF, and FPVDR datasets.}
\label{tab:quan_res}
\centering
\vspace{-0.25cm}
\setlength{\tabcolsep}{1.2mm}
\renewcommand{\arraystretch}{1.2}
\resizebox{\textwidth}{!}{
\begin{tabular}{l cccc cccc cccc crrr}
    \multirow{2}{*}{} &&
    \multicolumn{3}{c}{ECD} && 
    \multicolumn{3}{c}{MVSEC} && 
    \multicolumn{3}{c}{HQF} && 
    \multicolumn{1}{c}{FAST} &
    \multicolumn{1}{c}{NIGHT} &
    \multicolumn{1}{c}{FPVDR} 
    \\
    \cmidrule{3-5} \cmidrule{7-9} \cmidrule{11-13} \cmidrule{15-15} \cmidrule{16-16} \cmidrule{17-17} 
    && MSE $\downarrow$  & SSIM $\uparrow$  & LPIPS $\downarrow$ 
    && MSE $\downarrow$ & SSIM $\uparrow$ & LPIPS $\downarrow$
    && MSE $\downarrow$ & SSIM $\uparrow$ & LPIPS $\downarrow$
    & \multicolumn{4}{c}{BRISQUE $\downarrow$}  \\
    \midrule
    E2VID~\cite{rebecq2019high} && 
    0.179 & 0.450 & 0.322 && 0.225 & 0.241 & 0.644 && 0.098 & 0.468 & 0.371 && \underline{14.957} & ~\textbf{2.153} & ~\underline{14.239} \\
    FireNet~\cite{Scheerlinck20wacv} && 
    0.131 & 0.459 & 0.320 && 0.292 & 0.199 & 0.700 && 0.094 & 0.423 & 0.441 && 19.957 & 21.311 & 21.395 \\
    E2VID+~\cite{stoffregen2020reducing} && 
    0.070 & 0.503 & 0.236 && 0.132 & 0.262 & 0.514 && 0.036 & \underline{0.533} & \textbf{0.252} && 22.627 & 12.285 & 18.677 \\
    FireNet+~\cite{stoffregen2020reducing} && 
    0.063 & 0.452 & 0.290 && 0.218 & 0.212 & 0.570 && 0.040 & 0.471 & 0.314 && 18.399 & 10.019 & 15.502 \\
    SPADE-E2VID~\cite{cadena2021spade} && 
    0.091 & 0.461 & 0.337 && 0.138 & 0.266 & 0.589 && 0.077 & 0.400 & 0.502 && 18.925 & 24.011 & 21.248 \\
    SSL-E2VID~\cite{paredes2021back} && 
    0.092 & 0.415 & 0.380 && 0.124 & 0.264 & 0.693 && 0.082 & 0.421 & 0.467 && 46.199 & 49.562 & 59.454 \\
    ET-Net~\cite{weng2021event} && 
\underline{0.047} & \underline{0.552} & \underline{0.224} && \underline{0.107} & \underline{0.288} & \underline{0.489} && \underline{0.032} & \textbf{0.534} & 0.260 && 19.698 & 15.533 & 22.745 \\
    \textbf{HyperE2VID (ours)} && 
    \textbf{0.033} & \textbf{0.576} & \textbf{0.212} && \textbf{0.076} & \textbf{0.315} & \textbf{0.476} && \textbf{0.031} & 0.530 & \underline{0.257} && \textbf{14.024} & ~\underline{5.973} & ~\textbf{14.178}\\
    \bottomrule
\end{tabular}\vspace{-0.25cm}}
\end{table*}

%% file: tables/qual_eval.tex
\begin{figure*}[!t]
\small
	\newcommand{\widthplot}{0.132\textwidth}
	\centering
	\setlength{\tabcolsep}{0.3ex} %
\resizebox{0.95\textwidth}{!}{ 
\begin{tabular}{cccccccc}
    \rotatebox[origin=l]{90}{~~~~Boxes} &
    \includegraphics[width=\widthplot]{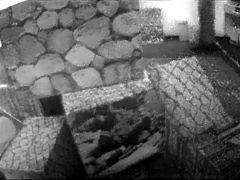} &
	\includegraphics[width=\widthplot]{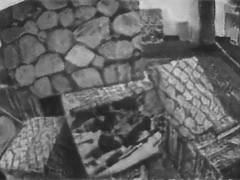} &
	\includegraphics[width=\widthplot]{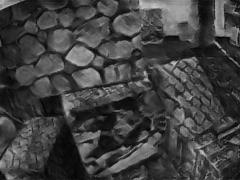} &
	\includegraphics[width=\widthplot]{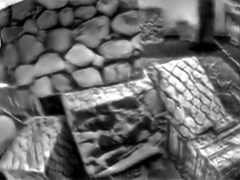} &
	\includegraphics[width=\widthplot]{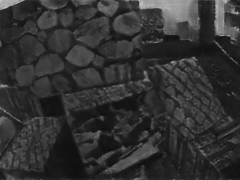} &
	\includegraphics[width=\widthplot]{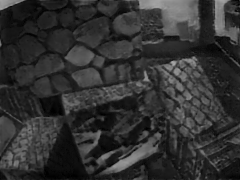} &
	\includegraphics[width=\widthplot]{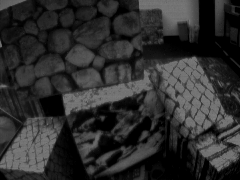} \\	
	
	\rotatebox[origin=l]{90}{~~~~Office} &
	\includegraphics[width=\widthplot]{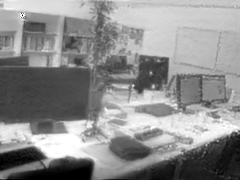} &
	\includegraphics[width=\widthplot]{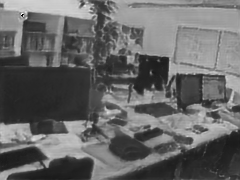} &
	\includegraphics[width=\widthplot]{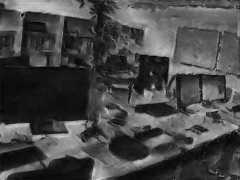} &
	\includegraphics[width=\widthplot]{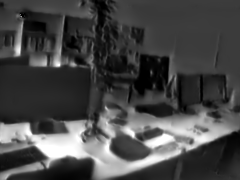} &
	\includegraphics[width=\widthplot]{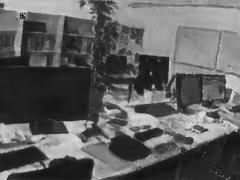} &
	\includegraphics[width=\widthplot]{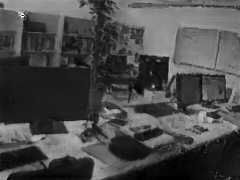} &
	\includegraphics[width=\widthplot]{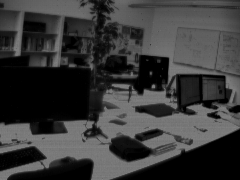} \\	
	
	\rotatebox[origin=l]{90}{~~In Flight} &
	\includegraphics[width=\widthplot]{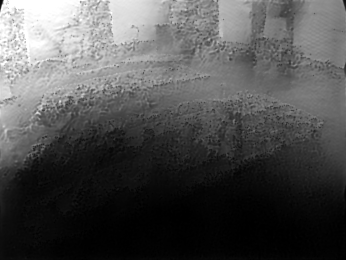} &
	\includegraphics[width=\widthplot]{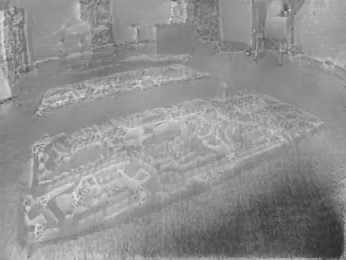} &
	\includegraphics[width=\widthplot]{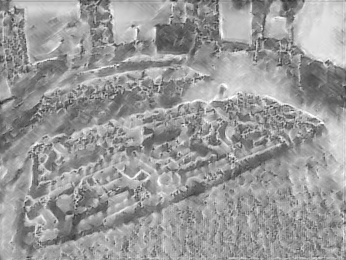} &
	\includegraphics[width=\widthplot]{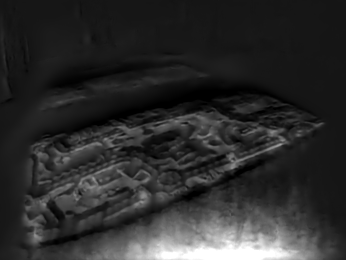} &
	\includegraphics[width=\widthplot]{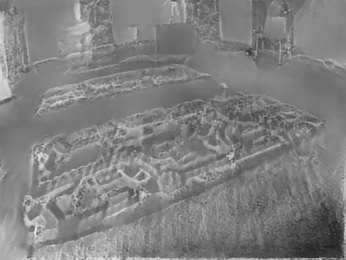} &
	\includegraphics[width=\widthplot]{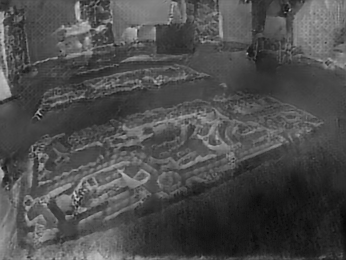} &
	\includegraphics[width=\widthplot]{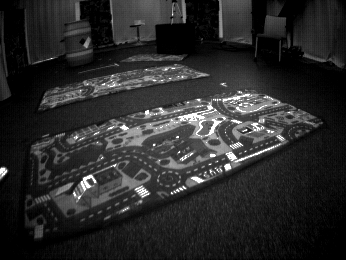} \\
	
	\rotatebox[origin=l]{90}{~~Out Day} &
	\includegraphics[width=\widthplot]{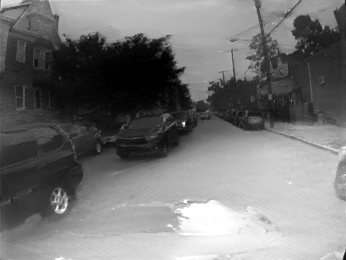} &
	\includegraphics[width=\widthplot]{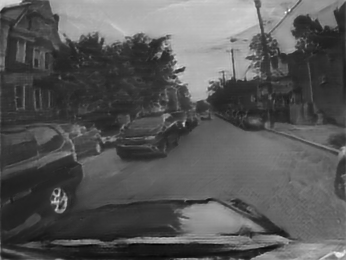} &
	\includegraphics[width=\widthplot]{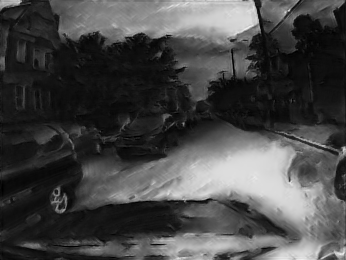} &
	\includegraphics[width=\widthplot]{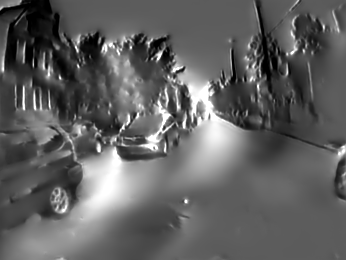} &
	\includegraphics[width=\widthplot]{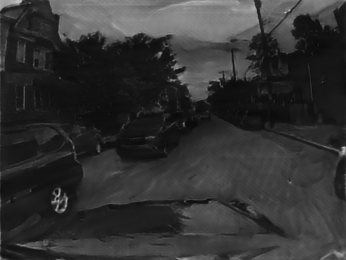} &
	\includegraphics[width=\widthplot]{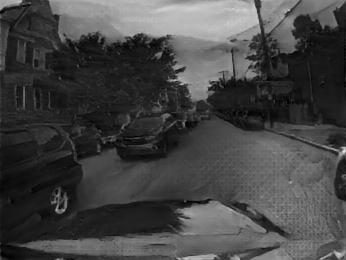} &
	\includegraphics[width=\widthplot]{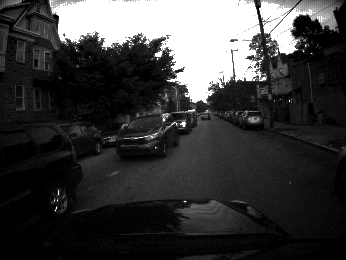} \\
	
	\rotatebox[origin=l]{90}{~Desk Fast} &
	\includegraphics[width=\widthplot]{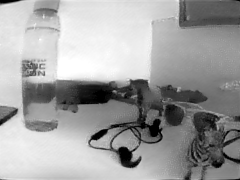} &
	\includegraphics[width=\widthplot]{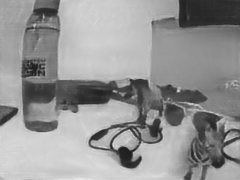} &
	\includegraphics[width=\widthplot]{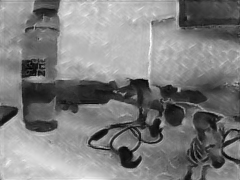} &
	\includegraphics[width=\widthplot]{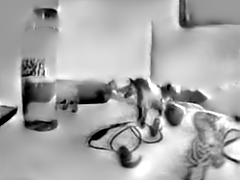} &
	\includegraphics[width=\widthplot]{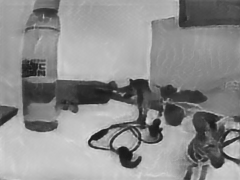} &
	\includegraphics[width=\widthplot]{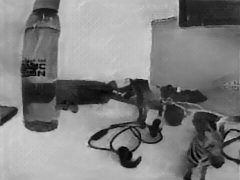} &
	\includegraphics[width=\widthplot]{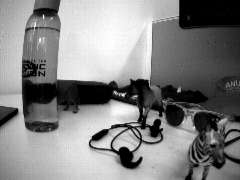} \\

	\rotatebox[origin=l]{90}{~Desk Slow} &
	\includegraphics[width=\widthplot]{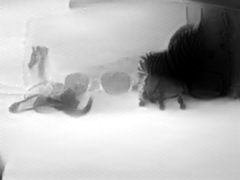} &
	\includegraphics[width=\widthplot]{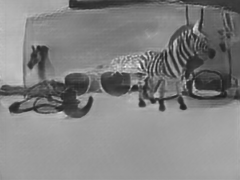} &
	\includegraphics[width=\widthplot]{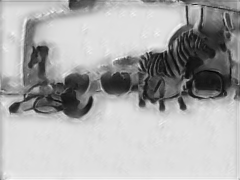} &
	\includegraphics[width=\widthplot]{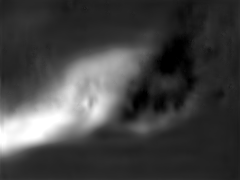} &
	\includegraphics[width=\widthplot]{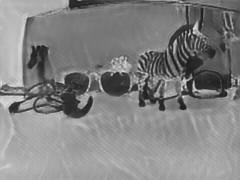} &
	\includegraphics[width=\widthplot]{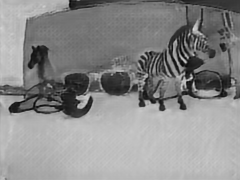} &
	\includegraphics[width=\widthplot]{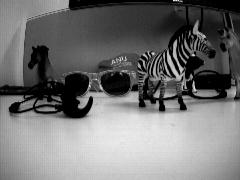} \\
 
	\rotatebox[origin=l]{90}{~~~Texture} &
	\includegraphics[width=\widthplot]{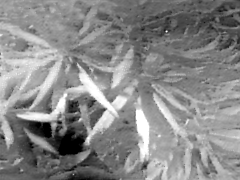} &
	\includegraphics[width=\widthplot]{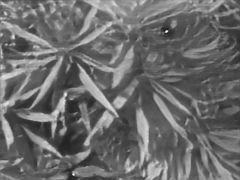} &
	\includegraphics[width=\widthplot]{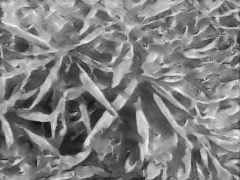} &
	\includegraphics[width=\widthplot]{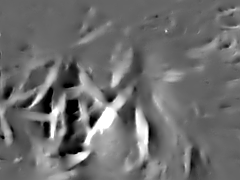} &
	\includegraphics[width=\widthplot]{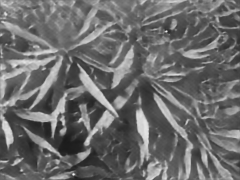} &
	\includegraphics[width=\widthplot]{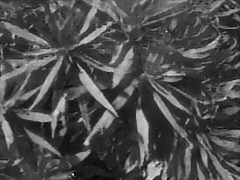} &
	\includegraphics[width=\widthplot]{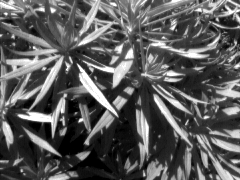} \\

	\rotatebox[origin=l]{90}{Drone Race} &
	\includegraphics[width=\widthplot]{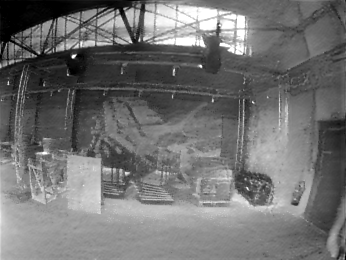} &
	\includegraphics[width=\widthplot]{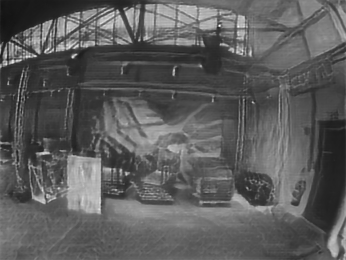} &
	\includegraphics[width=\widthplot]{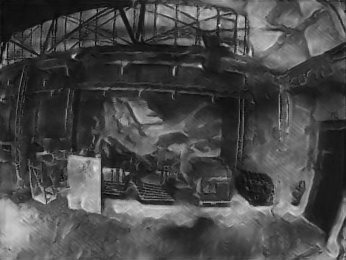} &
	\includegraphics[width=\widthplot]{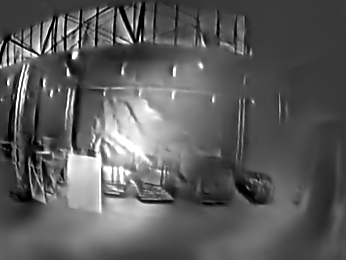} &
	\includegraphics[width=\widthplot]{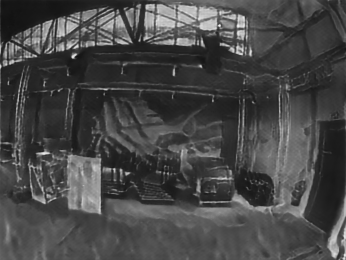} &
	\includegraphics[width=\widthplot]{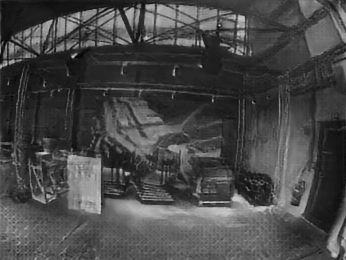} &
	\includegraphics[width=\widthplot]{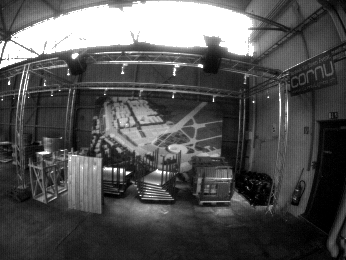} \\

	\rotatebox[origin=l]{90}{~~Dynamic} &
	\includegraphics[width=\widthplot]{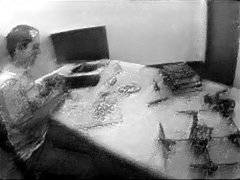} &
	\includegraphics[width=\widthplot]{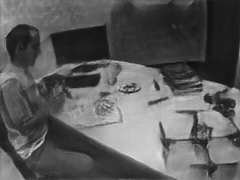} &
	\includegraphics[width=\widthplot]{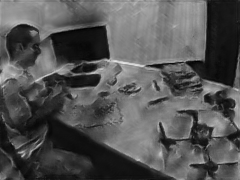} &
	\includegraphics[width=\widthplot]{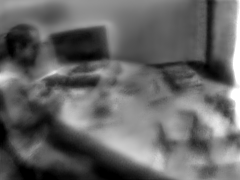} &
	\includegraphics[width=\widthplot]{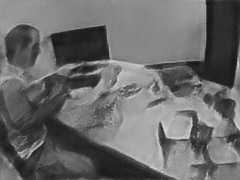} &
	\includegraphics[width=\widthplot]{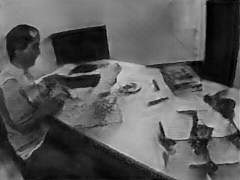} &
	\includegraphics[width=\widthplot]{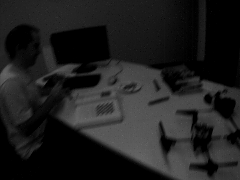} \\

	\rotatebox[origin=l]{90}{~~Out Night} &
	\includegraphics[width=\widthplot]{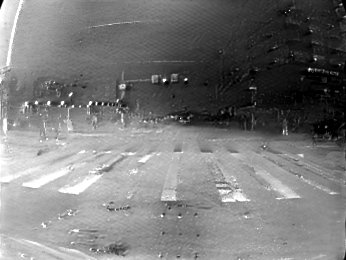} &
	\includegraphics[width=\widthplot]{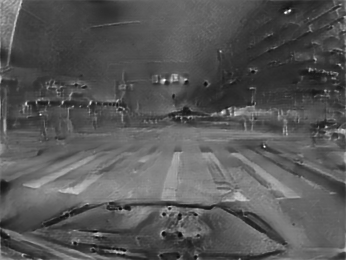} &
	\includegraphics[width=\widthplot]{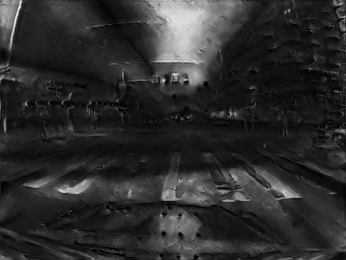} &
	\includegraphics[width=\widthplot]{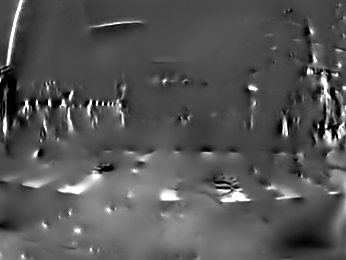} &
	\includegraphics[width=\widthplot]{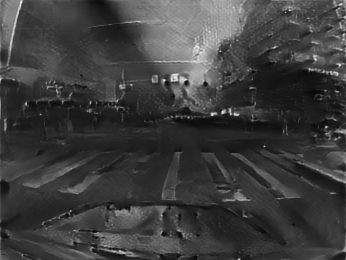} &
	\includegraphics[width=\widthplot]{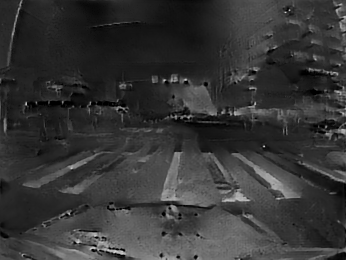} &
	\includegraphics[width=\widthplot]{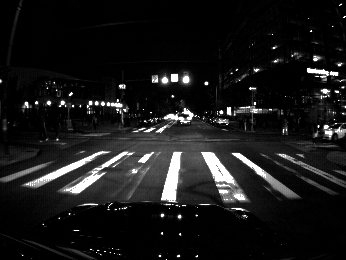} \\
 
\small
	 & E2VID & E2VID+ & FireNet+ & SSL-E2VID & ET-Net & HyperE2VID & Reference Frame
\end{tabular}}
	\caption{\textbf{Qualitative comparisons on some sequences from ECD (rows 1-2), MVSEC (rows 3-4), HQF (rows 5-7), FPVDR (row 8), FAST (row 9), and NIGHT (row 10).} While the competing approaches suffer from low contrast, blur, and extensive artifacts, HyperE2VID reconstructions have high contrast and preserve sharp details around the edges, with minimal artifacts in textureless regions.}
	\label{fig:qual_eval}
\end{figure*}

%% file: tables/color_recons.tex
\begin{figure}[!ht]
\small
\centering
\begin{minipage}[t]{\columnwidth}
\setlength{\tabcolsep}{0.2ex} %
\begin{tabular}{lcccc}

    \rotatebox[origin=l]{90}{~~Carpet} &
    \includegraphics[width=0.23\linewidth]{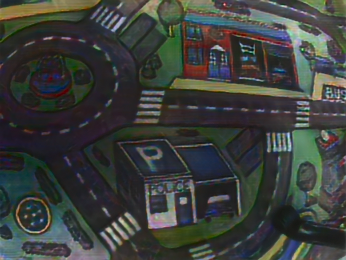} & 
    \includegraphics[width=0.23\linewidth]{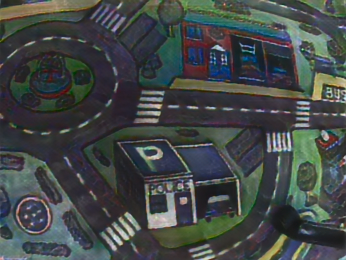} & 
    \includegraphics[width=0.23\linewidth]{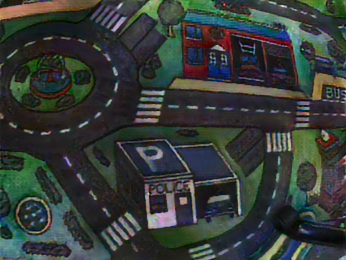} & 
    \includegraphics[width=0.23\linewidth]{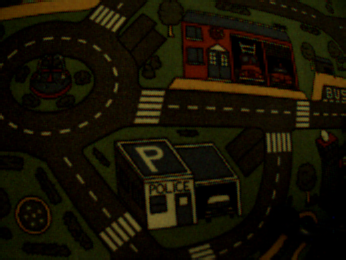} \\

    \rotatebox[origin=l]{90}{~~~Fruit} &
    \includegraphics[width=0.23\linewidth]{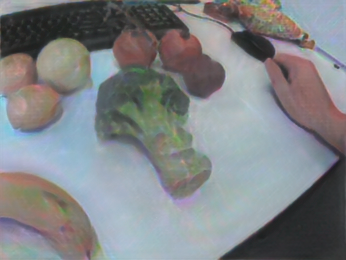} & 
    \includegraphics[width=0.23\linewidth]{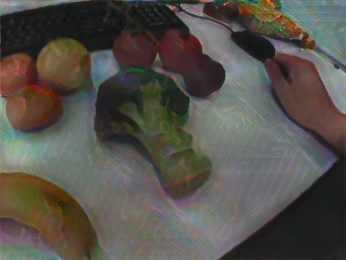} & 
    \includegraphics[width=0.23\linewidth]{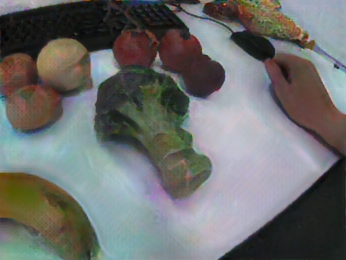} & 
    \includegraphics[width=0.23\linewidth]{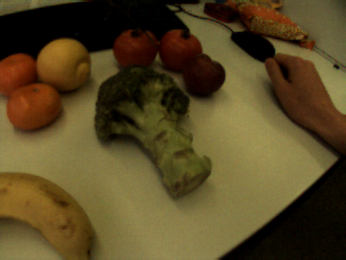} \\

    \rotatebox[origin=l]{90}{~~~Jenga} &
    \includegraphics[width=0.23\linewidth]{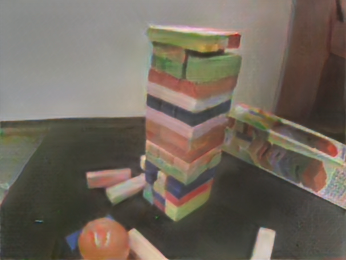} & 
    \includegraphics[width=0.23\linewidth]{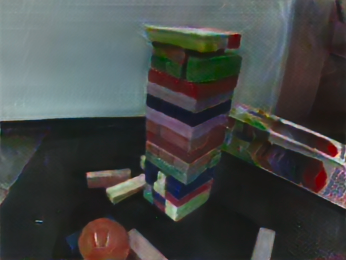} & 
    \includegraphics[width=0.23\linewidth]{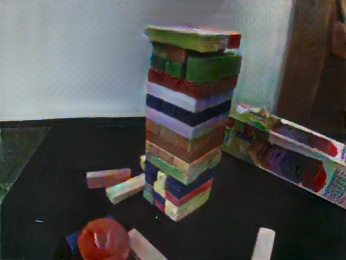} & 
    \includegraphics[width=0.23\linewidth]{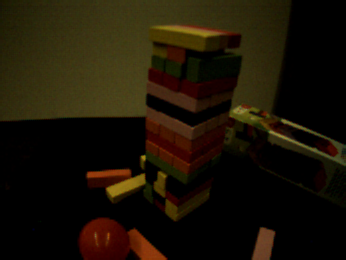} \\

    \rotatebox[origin=l]{90}{~~~HDR} &
    \includegraphics[width=0.23\linewidth]{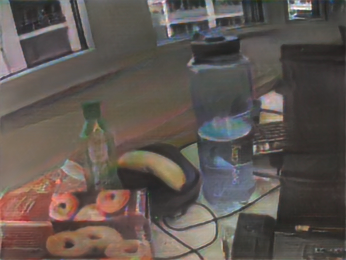} & 
    \includegraphics[width=0.23\linewidth]{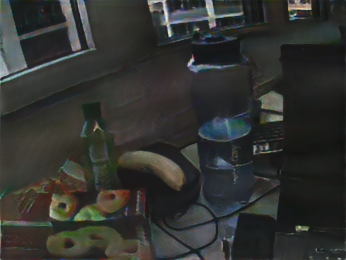} & 
    \includegraphics[width=0.23\linewidth]{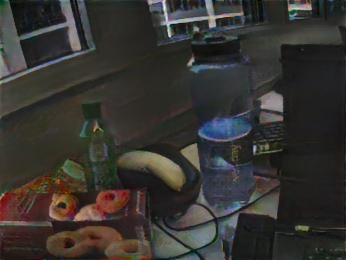} & 
    \includegraphics[width=0.23\linewidth]{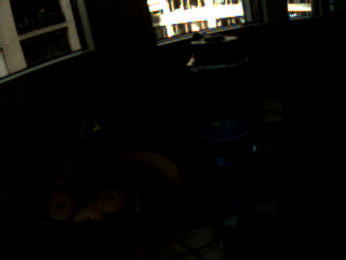} \\

    & E2VID+ & ET-Net & HyperE2VID & Reference \\    
	
\end{tabular}
\caption{\textbf{Color image reconstructions on CED.} HyperE2VID excels in reconstructing visually appealing scenes from the CED dataset, including those with colorful objects and HDR scenarios, outperforming E2VID+ and ET-Net in visual quality.}
\label{fig:color}
\end{minipage}
\end{figure}

%% file: chapters/04F_experiments_comp2.tex
\subsection{Computational Complexity} \label{sec:experiments_comp2}

We also analyze the computational complexity of our method and compare it to other competing methods from the literature. We consider three computational metrics for this analysis: (1) the number of model parameters, (2) the number of floating point operations (FLOPs), and (3) inference time. The number of parameters is an important metric that indicates the memory requirements of the model, while FLOPs specify the computational requirements and efficiency, and finally, the inference time is a direct indicator of the real-time performance of (the maximum frame-per-seconds that can be obtained with) the model. We use data with a resolution of 240$ \times $180 to measure FLOPs and inference time, where the average inference times are calculated on a workstation with Quadro RTX 5000 GPU. We present the results of these computational complexity metrics in Table~\ref{tab:comp}. Here, the numbers of model parameters are given in millions, FLOPs are given in billions (as GFLOPs), and the inference times are given in milliseconds. Methods that share a common network architecture are presented in the same row. Here, it can be seen that our method provides a good trade-off between accuracy and efficiency. HyperE2VID is a significantly smaller and faster network than ET-Net while generating reconstructions with better visual quality. On the other hand, the smallest and fastest methods, FireNet and FireNet+, generate reconstructions with significantly lower visual quality.
\input{tables/computational.tex}

%% file: tables/computational.tex
\begin{table}[!t]
\small
\begin{minipage}[t]{\columnwidth}
\caption{Computational complexity of network architectures in terms of the number of model parameters (in millions), number of floating point operations (FLOPS - in billions), and inference time (in milliseconds).}

\label{tab:comp}
\centering
\centering
\setlength{\tabcolsep}{1.2mm}
\renewcommand{\arraystretch}{1.2} 
\begin{tabular}{lr rc rc r}
    \multicolumn{1}{p{2.4cm}}{Network \newline Architecture}  && 
    \multicolumn{1}{p{1.5cm}}{Number of \newline Params (M)} && 
    \multicolumn{1}{p{1.2cm}}{~ \newline GFLOPs} && 
    \multicolumn{1}{p{1.4cm}}{Inference \newline Time (ms)} \\ 
    \midrule
    E2VID~\cite{rebecq2019high,stoffregen2020reducing,paredes2021back} && 
    10.71 && 20.07 && ~{\underline{5.1}} \\
    FireNet~\cite{Scheerlinck20wacv,stoffregen2020reducing}  && 
    ~\textbf{0.04}	&& \textbf{1.62}	&& ~\textbf{1.6}  \\
    SPADE-E2VID~\cite{cadena2021spade} && 
    11.46	&& 68.06 && 16.1 \\
    ET-Net~\cite{weng2021event} && 
    22.18	&& 33.10	&& 32.1 \\
    \textbf{HyperE2VID (ours)} && 
    \underline{10.15}	&& \underline{18.46}	&& ~6.6 \\
    \bottomrule
\end{tabular}
\end{minipage}
\end{table}

%% file: chapters/05_conclusion.tex
\section{Conclusion} \label{sec:conclusion}
In this work, we present HyperE2VID, a novel dynamic network architecture for event-based video reconstruction that improves the state-of-the-art by employing hypernetworks and dynamic convolutions. Our approach generates adaptive filters using hypernetworks, which are dynamically generated at inference time based on the scene context encoded via event voxel grids and previously reconstructed intensity images, and thus deals with static and dynamic parts of the scene more effectively. Experimental results on several challenging datasets show that HyperE2VID outperforms previous state-of-the-art methods in terms of visual quality while reducing memory consumption, FLOPs, and inference time. Our work demonstrates the potential of dynamic network architectures and hypernetworks for processing highly varying event data, opening up possibilities for future research in this direction, targeting more tasks like event-based optical-flow estimation.

%% file: tables_supp/ablation_hyper.tex
\begin{table*}[!t]
\small
\caption{Results from ablation experiments investigating effects of training settings, use of previous reconstructions, dynamic convolutions, and hypernetworks.}
\label{tab:ablation_hyper}
\centering
\setlength{\tabcolsep}{1.2mm}
\renewcommand{\arraystretch}{1.2}
\begin{tabular}{l cccc cccc cccc crrr}
    \multirow{2}{*}{} &&
    \multicolumn{3}{c}{ECD} && 
    \multicolumn{3}{c}{MVSEC} && 
    \multicolumn{3}{c}{HQF} && 
    \multicolumn{1}{c}{FAST} &
    \multicolumn{1}{c}{NIGHT} &
    \multicolumn{1}{c}{FPVDR} \\
    \cmidrule{3-5} \cmidrule{7-9} \cmidrule{11-13} \cmidrule{15-17}
    && MSE  & SSIM  & LPIPS  
    && MSE  & SSIM  & LPIPS  
    && MSE  & SSIM  & LPIPS 
    && \multicolumn{3}{c}{BRISQUE }  \\
    \midrule
    E2VID+ && 
    0.070 & 0.503 & 0.236 && 0.132 & 0.262 & 0.514 && 0.036 & \textbf{0.533} & \textbf{0.252} && 22.627 & 12.285 & 18.677 \\
    E2VID+ (re-trained) && 
    0.047 & 0.537 & \underline{0.217} && 0.153 & 0.259 & 0.531 && 0.048 & 0.507 & 0.285 && 17.719 & 8.131 & 15.432 \\  \midrule 
    w/ $\hat{I}_{k-1}$ at input && 
    0.077 & 0.479 & 0.259 && 0.226 & 0.218 & 0.567 && 0.037 & 0.496 & 0.270 && 18.830 & 10.983 & 21.176\\ 
    w/ Dynamic Conv.~\cite{chen2020dynamic} && 
    0.060 & 0.503 & 0.246 && \underline{0.119} & 0.270 & \underline{0.493} && \textbf{0.031} & 0.529 & \textbf{0.252} && \underline{15.162} & \underline{6.602} & 20.758 \\
    w/ CondConv~\cite{yang2019condconv} && 
    \underline{0.044} & \underline{0.565} & 0.221 && 0.119 & \underline{0.271} & 0.504 && 0.033 & 0.529 & 0.254 && 15.543 & 6.670 & \underline{14.264} \\
    HyperE2VID && 
    \textbf{0.033} & \textbf{0.576} & \textbf{0.212} && \textbf{0.076} & \textbf{0.315} & \textbf{0.476} && \textbf{0.031} & \underline{0.530} & 0.257  && \textbf{14.024} & \textbf{5.973} & \textbf{14.178} \\
    \bottomrule
\end{tabular}
\end{table*}

%% file: tables_supp/ablation_context_combined.tex
\begin{table*}[!t]
\small
\caption{Ablation results of HyperE2VID variants where we alter the context information, the existence of convolutional context fusion (CF) block, and curriculum learning (CL) strategy.}
\label{tab:ablation_context}
\centering
\setlength{\tabcolsep}{0.9mm}
\renewcommand{\arraystretch}{1.2}
\begin{tabular}{lcc cccc cccc cccc cccc cccc}
    &&&& 
    \multicolumn{3}{c}{ECD} && 
    \multicolumn{3}{c}{MVSEC} && 
    \multicolumn{3}{c}{HQF} && 
    \multicolumn{3}{c}{SLOW} && 
    \multicolumn{1}{c}{FAST} &
    \multicolumn{1}{c}{NIGHT} &
    \multicolumn{1}{c}{FPVDR}  \\
    \cmidrule{5-7} \cmidrule{9-11} \cmidrule{13-15} \cmidrule{17-19} \cmidrule{21-23}
    Context & CL & CF 
    && MSE  & SSIM  & LPIPS  
    && MSE  & SSIM  & LPIPS  
    && MSE  & SSIM  & LPIPS 
    && MSE  & SSIM  & LPIPS 
    && \multicolumn{3}{c}{BRISQUE}
     \\
    \midrule
    EVG &&&& 
    0.048 & 0.543 & 0.219 && 0.189 & 0.232 & 0.549 && 0.050 & 0.504 & 0.280 && 
    0.064 & 0.496 & 0.333 && \textbf{13.42} & ~\textbf{1.05} & ~\textbf{4.66} \\
    PR &&&& 
    0.050 & 0.536 & 0.229 && 0.181 & 0.228 & 0.573 && 0.035 & 0.517 & 0.276 && 
    {\underline{0.039}} & 0.558 & 0.283 && 18.79 & 10.12 & 16.18 \\
    EVG+PR &&&& 
    0.039 & {\underline{0.559}} & \textbf{0.212} && 0.152 & 0.261 & 0.532 && 0.036 & 0.525 & 0.271 && 
    0.045 & 0.566 & 0.279 && 18.64 & ~9.21 & 14.44 \\
    EVG+PR &&  \checkmark && 
    0.044 & 0.548 & 0.218 && \underline{0.113} & {\underline{0.274}} & 0.516 && 0.039 & 0.520 & 0.266 && 
    0.044 & {\underline{0.569}} & {\underline{0.268}} && 17.73 & ~7.38 & \underline{12.46} \\
    EVG+PR &  \checkmark &&& 
    \underline{0.038} & 0.556 & 0.216 && 0.120 & 0.265 & \underline{0.506} && \underline{0.032} & \textbf{0.534} & \underline{0.259} && 
    {\underline{0.039}} & 0.541 & 0.285 && 18.24 & ~6.09 & 13.26 \\
    EVG+PR & \checkmark & \checkmark && 
    \textbf{0.033} & \textbf{0.576} & \textbf{0.212} && \textbf{0.076} & \textbf{0.315} & \textbf{0.476} && \textbf{0.031} & {\underline{0.530}} & \textbf{0.257}  && 
    \textbf{0.026} & \textbf{0.581} & \textbf{0.250} && \underline{14.02} & ~\underline{5.97} & 14.18 \\
    \bottomrule
\end{tabular}
\end{table*}

%% file: figures_supp/slow_fast_im_ev_qual.tex
\begin{figure*}[!b]
\small
\centering
\setlength{\tabcolsep}{0.4ex} %
\begin{tabular}{lccccc}

    \rotatebox[origin=l]{90}{\quad Fast Motion} &
    \includegraphics[width=0.18\linewidth]{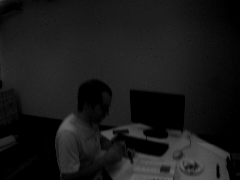} & 
    \includegraphics[width=0.18\linewidth]{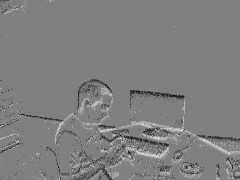} & 
    \includegraphics[width=0.18\linewidth]{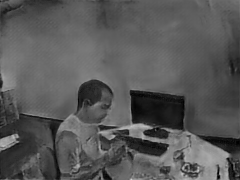} & 
    \includegraphics[width=0.18\linewidth]{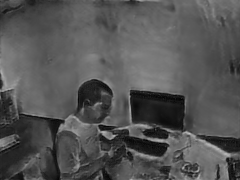} & 
    \includegraphics[width=0.18\linewidth]{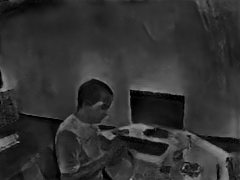} \\

    \rotatebox[origin=l]{90}{\quad Slow Motion} &
    \includegraphics[width=0.18\linewidth]{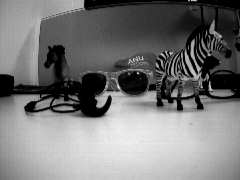} & 
    \includegraphics[width=0.18\linewidth]{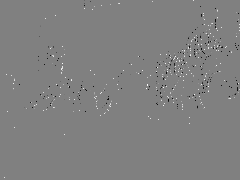} & 
    \includegraphics[width=0.18\linewidth]{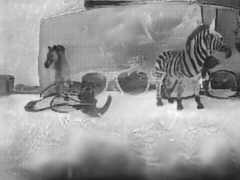} & 
    \includegraphics[width=0.18\linewidth]{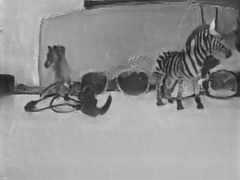} & 
    \includegraphics[width=0.18\linewidth]{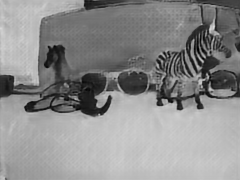} \\

    & Frames & Events & EVG & PR & HyperE2VID \\    
	
\end{tabular}
\caption{\textbf{Understanding the role of context information.} This figure shows frames, events, and reconstructions from two distinct scenes: one with fast motion (top) and another with slow motion (bottom). It highlights the significance of utilizing event and reconstruction data as context information for optimal results. }
\label{fig:sfieq}
\end{figure*}

%% file: figures_supp/robustness_plot.tex
\begin{figure*}
  \small
  \centering
  \subfloat[Different temporal windows]{\includegraphics[width=0.4\textwidth]{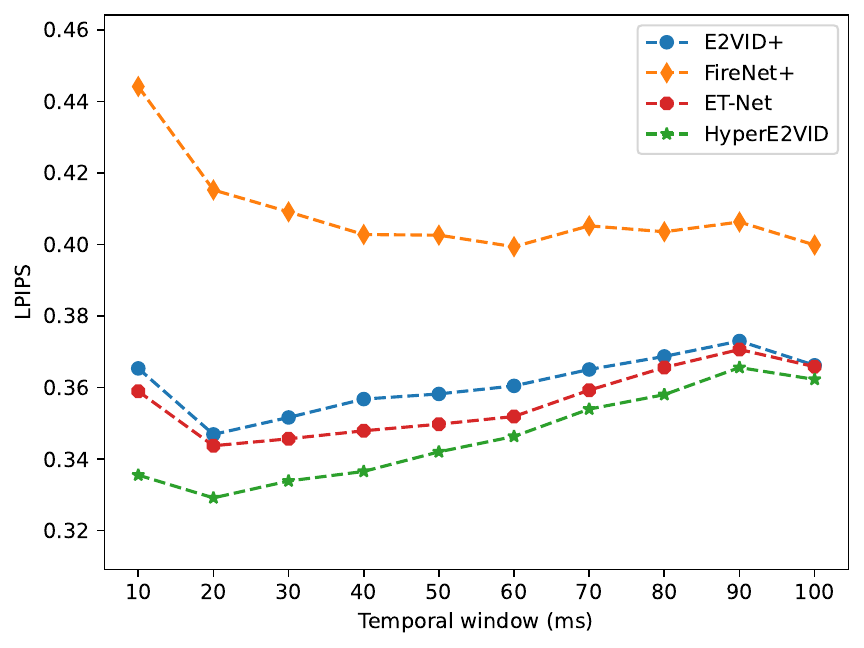}} \hspace{1cm}
  \subfloat[Different event numbers]{\includegraphics[width=0.4\textwidth]{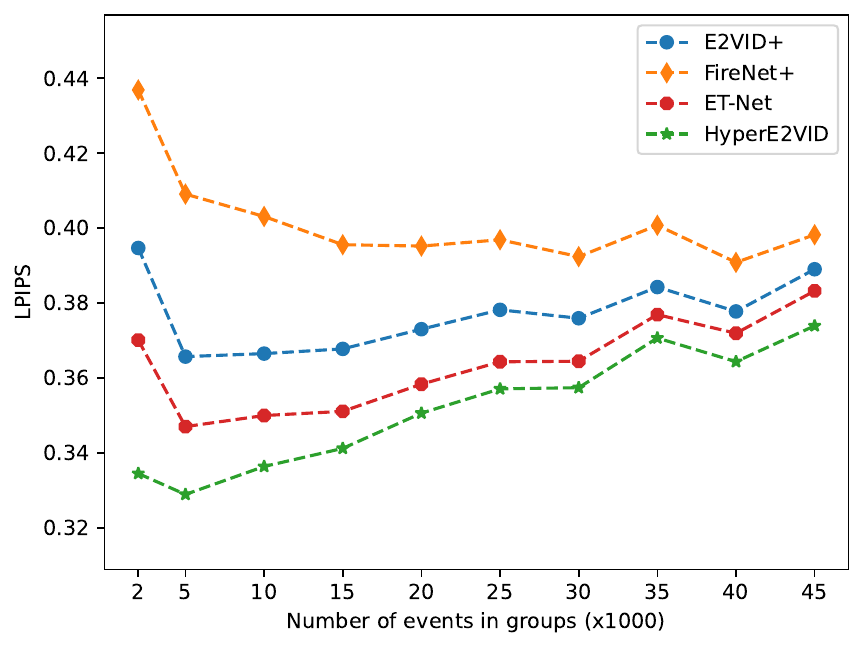}}
  \caption{\textbf{Effect of using event voxel grids with different temporal windows and event numbers.} We consider four best performing methods (E2VID+, FireNet+, ET-Net, and HyperE2VID), and compute their mean LPIPS scores obtained on ECD, MVSEC, and HQF datasets, using a variety of event grouping settings. (a) We conduct ten sets of experiments, each using a different temporal window ranging from \SI{10}{\milli\second} to \SI{100}{\milli\second}. (b) We conduct ten experiment runs, each utilizing fixed-number event grouping with a different event count ranging from 2K to 45K. For (a) and (b), we employ a tolerance of 1 ms to match the reconstructions with ground truth frames, and calculate LPIPS scores whenever there is a match. Then, we plot mean LPIPS scores across these experiments runs for each method. The results demonstrate the superiority of the proposed HyperE2VID architecture for generating high-quality reconstructions, over a wide range of event grouping settings.}
  \label{fig:robustness_plots}
\end{figure*}

%% file: figures_supp/high_FPS_recons.tex
\begin{figure*}[!b]
\small
\centering
\setlength{\tabcolsep}{0.4ex} %
\begin{tabular}{lccccc}

    \rotatebox[origin=l]{90}{\quad \quad E2VID} &
    \includegraphics[width=0.18\linewidth]{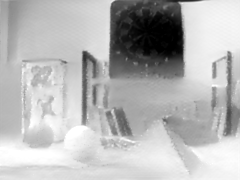} & 
    \includegraphics[width=0.18\linewidth]{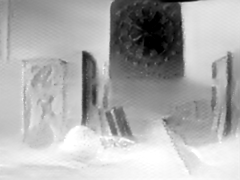} & 
    \includegraphics[width=0.18\linewidth]{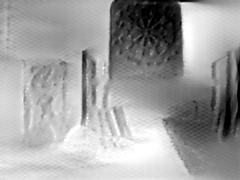} & 
    \includegraphics[width=0.18\linewidth]{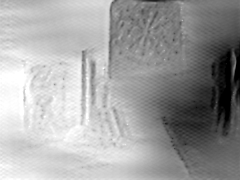} & 
    \includegraphics[width=0.18\linewidth]{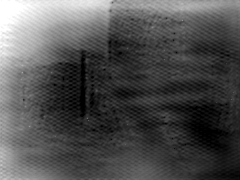} \\

    \rotatebox[origin=l]{90}{\quad \quad FireNet} &
    \includegraphics[width=0.18\linewidth]{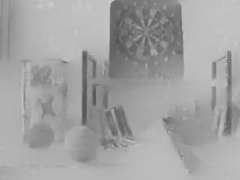} & 
    \includegraphics[width=0.18\linewidth]{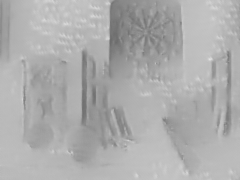} & 
    \includegraphics[width=0.18\linewidth]{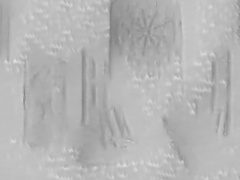} & 
    \includegraphics[width=0.18\linewidth]{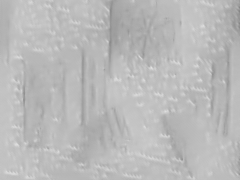} & 
    \includegraphics[width=0.18\linewidth]{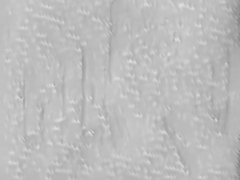} \\

    \rotatebox[origin=l]{90}{\quad \quad E2VID+} &
    \includegraphics[width=0.18\linewidth]{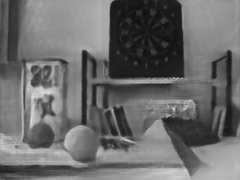} & 
    \includegraphics[width=0.18\linewidth]{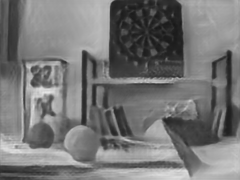} & 
    \includegraphics[width=0.18\linewidth]{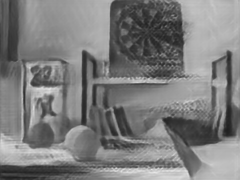} & 
    \includegraphics[width=0.18\linewidth]{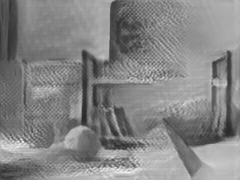} & 
    \includegraphics[width=0.18\linewidth]{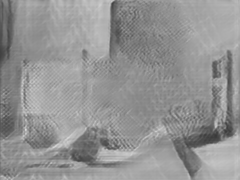} \\

    \rotatebox[origin=l]{90}{\quad ~~~ FireNet+} &
    \includegraphics[width=0.18\linewidth]{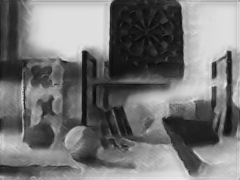} & 
    \includegraphics[width=0.18\linewidth]{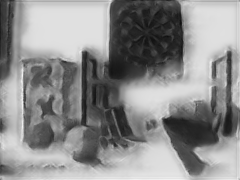} & 
    \includegraphics[width=0.18\linewidth]{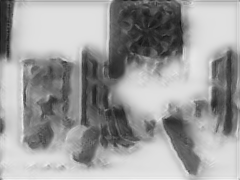} & 
    \includegraphics[width=0.18\linewidth]{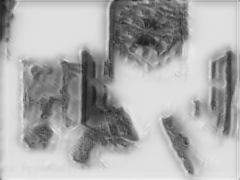} & 
    \includegraphics[width=0.18\linewidth]{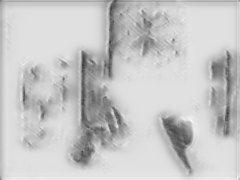} \\

    \rotatebox[origin=l]{90}{ SPADE-E2VID} &
    \includegraphics[width=0.18\linewidth]{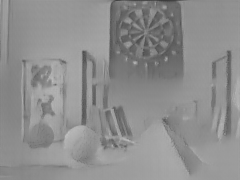} & 
    \includegraphics[width=0.18\linewidth]{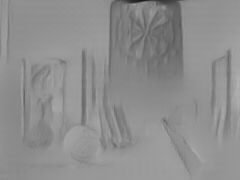} & 
    \includegraphics[width=0.18\linewidth]{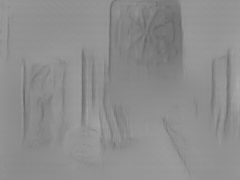} & 
    \includegraphics[width=0.18\linewidth]{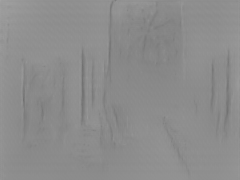} & 
    \includegraphics[width=0.18\linewidth]{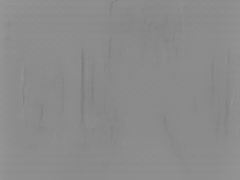} \\

    \rotatebox[origin=l]{90}{\quad SSL-E2VID} &
    \includegraphics[width=0.18\linewidth]{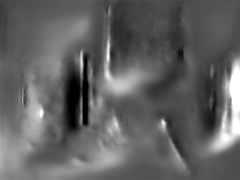} & 
    \includegraphics[width=0.18\linewidth]{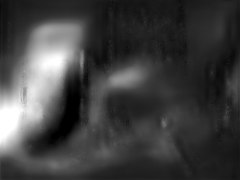} & 
    \includegraphics[width=0.18\linewidth]{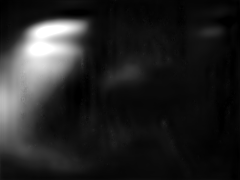} & 
    \includegraphics[width=0.18\linewidth]{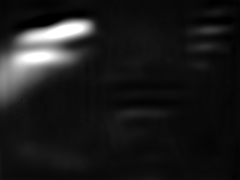} & 
    \includegraphics[width=0.18\linewidth]{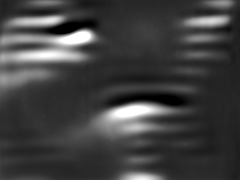} \\

    \rotatebox[origin=l]{90}{\quad \quad ET-Net} &
    \includegraphics[width=0.18\linewidth]{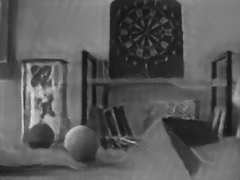} & 
    \includegraphics[width=0.18\linewidth]{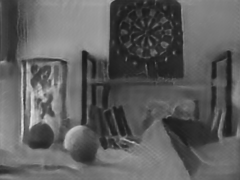} & 
    \includegraphics[width=0.18\linewidth]{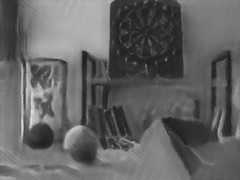} & 
    \includegraphics[width=0.18\linewidth]{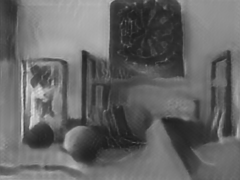} & 
    \includegraphics[width=0.18\linewidth]{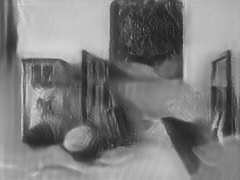} \\

    \rotatebox[origin=l]{90}{\quad HyperE2VID} &
    \includegraphics[width=0.18\linewidth]{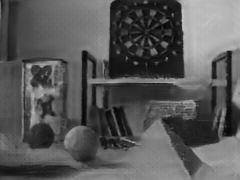} & 
    \includegraphics[width=0.18\linewidth]{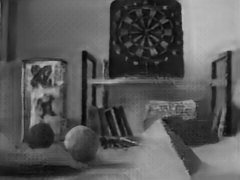} & 
    \includegraphics[width=0.18\linewidth]{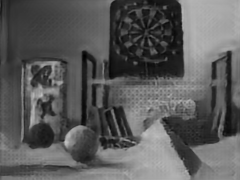} & 
    \includegraphics[width=0.18\linewidth]{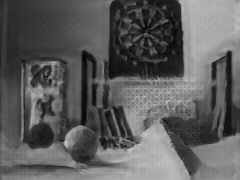} & 
    \includegraphics[width=0.18\linewidth]{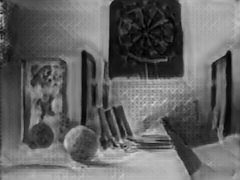} \\

    & 200 FPS & 500 FPS & 1000 FPS & 2000 FPS & 5000 FPS \\    
	
\end{tabular}
\caption{\textbf{High frame rate video synthesis.} We employ a simple approach with fixed-temporal-window event grouping for generating videos with high FPS. Here we present frames corresponding to the first second of the \texttt{slider\_depth} sequence from the ECD dataset, taken from videos reconstructed at \SI{200}{\hertz}, \SI{500}{\hertz}, \SI{1}{\kilo\hertz}, \SI{2}{\kilo\hertz}, and \SI{5}{\kilo\hertz}, which are generated by using temporal windows of \SI{5}{\milli\second}, \SI{2}{\milli\second}, \SI{1}{\milli\second}, \SI{500}{\micro\second}, and \SI{200}{\micro\second}, respectively. While most of the other methods start to generate videos with lower visual quality as we increase FPS above one thousand, HyperE2VID maintains its high contrast and sharp reconstructions even when generating videos with several thousand frames per second.}
\label{fig:high_fps}
\end{figure*}

%% file: figures_supp/pause_recons.tex
\begin{figure*}[!t]
\small
\newcommand{\widthplot}{0.18\textwidth}
\centering
\setlength{\tabcolsep}{0.6ex} %
\begin{tabular}{ccccc}
 
\rotatebox[origin=l]{90}{\quad \quad E2VID} &
\includegraphics[width=\widthplot]{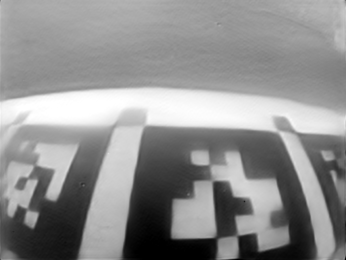} &
\includegraphics[width=\widthplot]{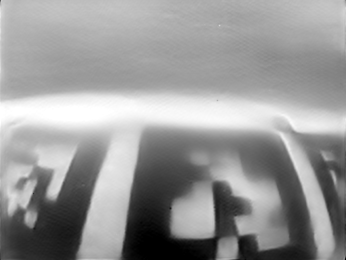} &
\includegraphics[width=\widthplot]{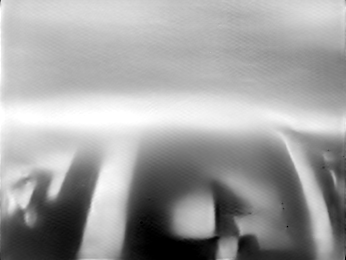} &
\includegraphics[width=\widthplot]{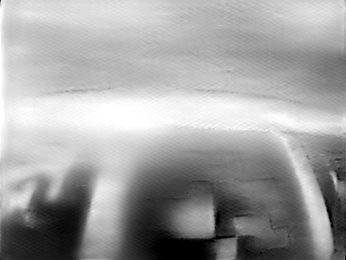} \\

\rotatebox[origin=l]{90}{\quad \quad FireNet} &
\includegraphics[width=\widthplot]{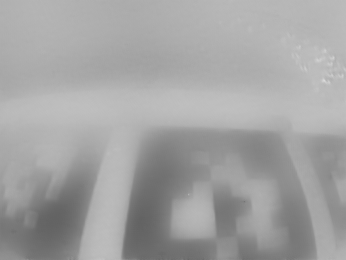} &
\includegraphics[width=\widthplot]{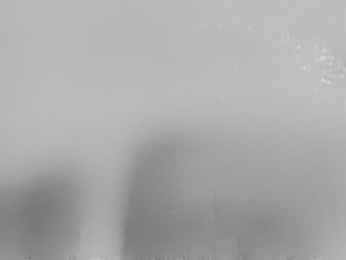} &
\includegraphics[width=\widthplot]{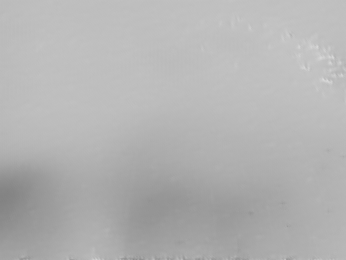} &
\includegraphics[width=\widthplot]{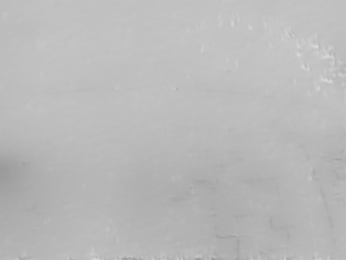} \\

 \rotatebox[origin=l]{90}{\quad \quad E2VID+} &
\includegraphics[width=\widthplot]{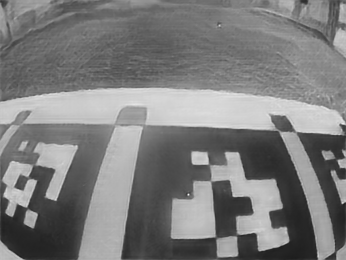} &
\includegraphics[width=\widthplot]{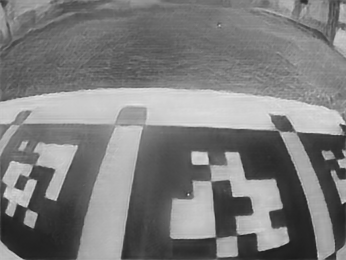} &
\includegraphics[width=\widthplot]{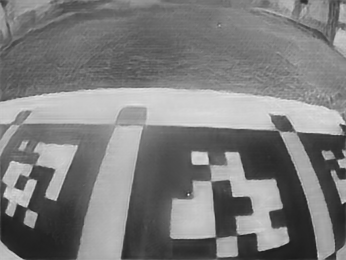} &
\includegraphics[width=\widthplot]{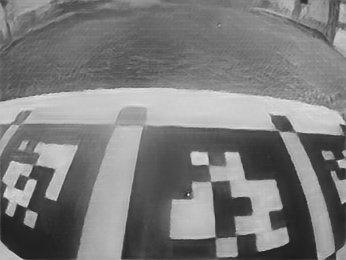} \\

\rotatebox[origin=l]{90}{\quad \quad FireNet+} &
\includegraphics[width=\widthplot]{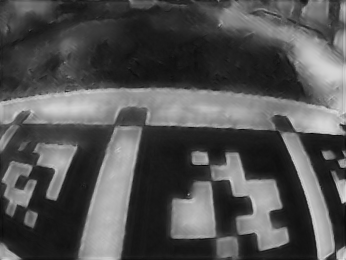} &
\includegraphics[width=\widthplot]{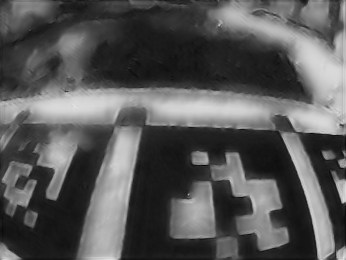} &
\includegraphics[width=\widthplot]{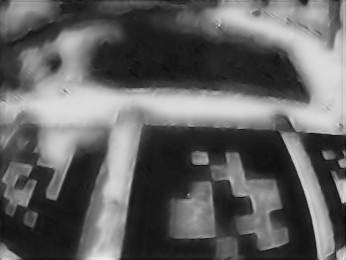} &
\includegraphics[width=\widthplot]{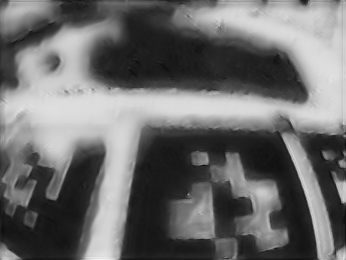} \\

 \rotatebox[origin=l]{90}{~~~ SPADE-E2VID} &
\includegraphics[width=\widthplot]{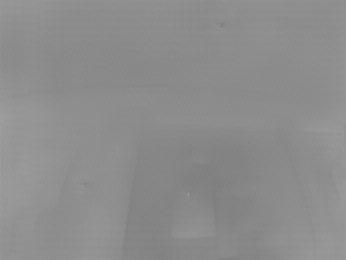} &
\includegraphics[width=\widthplot]{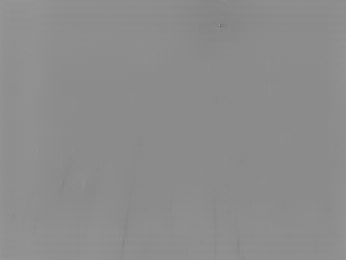} &
\includegraphics[width=\widthplot]{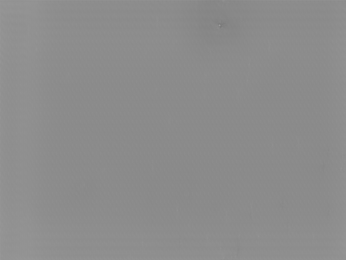} &
\includegraphics[width=\widthplot]{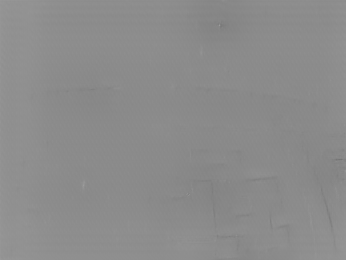} \\

 \rotatebox[origin=l]{90}{\quad SSL-E2VID} &
\includegraphics[width=\widthplot]{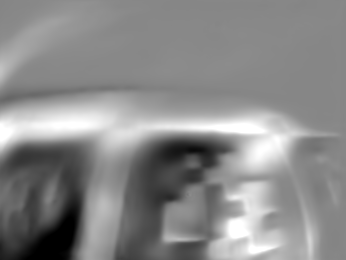} &
\includegraphics[width=\widthplot]{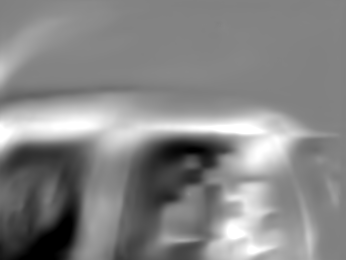} &
\includegraphics[width=\widthplot]{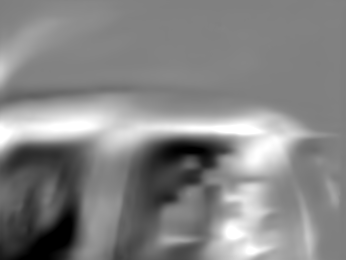} &
\includegraphics[width=\widthplot]{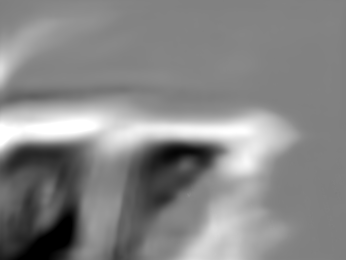} \\

 \rotatebox[origin=l]{90}{\quad \quad ET-Net} &
\includegraphics[width=\widthplot]{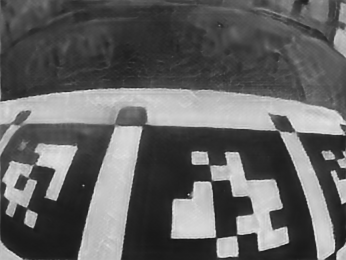} &
\includegraphics[width=\widthplot]{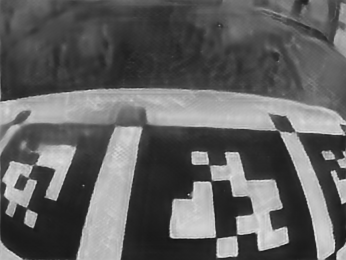} &
\includegraphics[width=\widthplot]{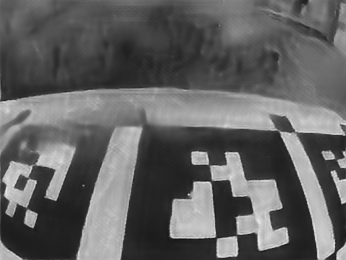} &
\includegraphics[width=\widthplot]{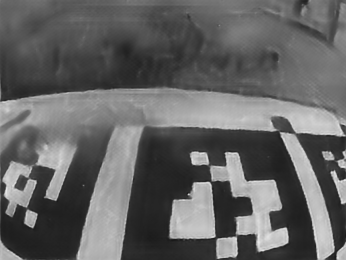} \\

 \rotatebox[origin=l]{90}{\quad HyperE2VID} &
\includegraphics[width=\widthplot]{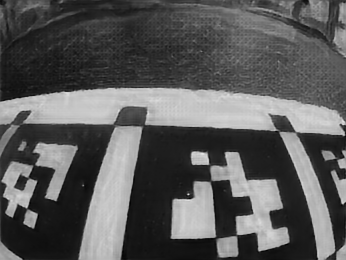} &
\includegraphics[width=\widthplot]{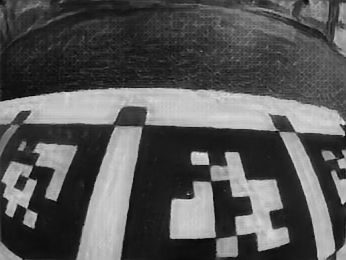} &
\includegraphics[width=\widthplot]{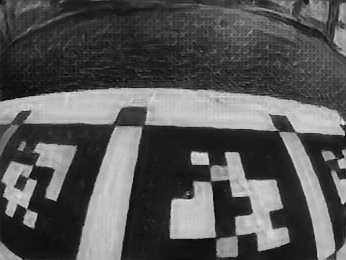} &
\includegraphics[width=\widthplot]{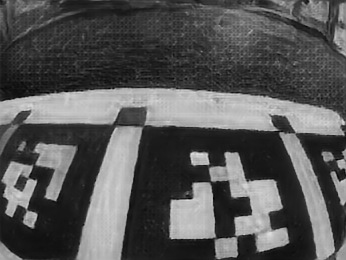} \\

& Initial Time & After 1 second & After 2 seconds & After 3 seconds 
 
\end{tabular}
\caption{\textbf{Assessing reconstruction quality in motionless sections.} Stationary sections in event sequences pose additional challenges for video reconstruction since the event rate drop-offs to almost zero, with only noise events being generated. Here, we consider a segment from the UZH-FPV Drone Racing dataset, where the drone lands on a board with ArUco markers and stops. For each method, we present reconstructions from the initial time just after the drone stops in the leftmost column and three more reconstructions at one-second intervals in subsequent columns. The desired functionality for methods is to retain their most recent reconstructions during the pause segment, but most of them start to generate intensity images with degraded quality within a few seconds by gradually decaying images and revealing artifacts such as blurry and bleeding edges. On the other hand, HyperE2VID manages to preserve its high contrast and sharp reconstructions during the motionless segments, thanks to its network architecture, which allows it to dynamically adapt to highly varying event data.}
\label{fig:pause}
\end{figure*}

%% file: figures_supp/post_process.tex
\begin{figure*}[!b]
\small
\centering
\renewcommand{\arraystretch}{1.2}
\setlength{\tabcolsep}{0.4ex} %
\begin{tabular}{lccccc}

    & \multicolumn{2}{c}{ECD - slider\_depth} 
    && \multicolumn{2}{c}{HQF - desk\_slow} \\

    \rotatebox[origin=l]{90}{\quad\quad~~ E2VID+} &
    \includegraphics[width=0.23\linewidth]{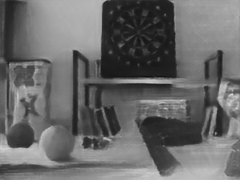} & 
    \includegraphics[width=0.23\linewidth]{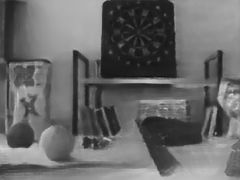} && 
    \includegraphics[width=0.23\linewidth]{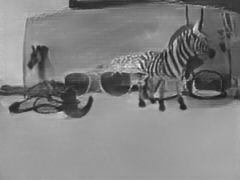} & 
    \includegraphics[width=0.23\linewidth]{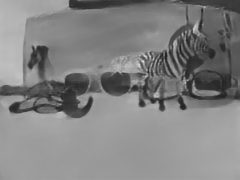} \\

    \rotatebox[origin=l]{90}{\quad\quad\quad ET-Net} &
    \includegraphics[width=0.23\linewidth]{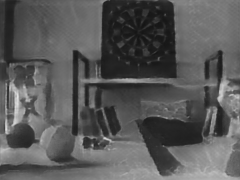} & 
    \includegraphics[width=0.23\linewidth]{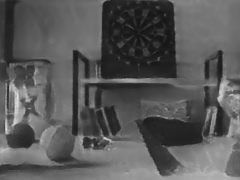} && 
    \includegraphics[width=0.23\linewidth]{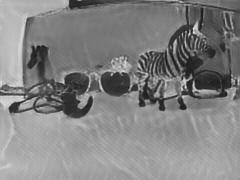} & 
    \includegraphics[width=0.23\linewidth]{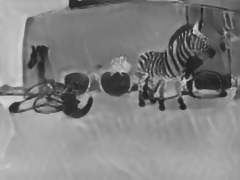} \\

    \rotatebox[origin=l]{90}{\quad\quad HyperE2VID} &
    \includegraphics[width=0.23\linewidth]{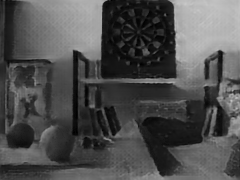} & 
    \includegraphics[width=0.23\linewidth]{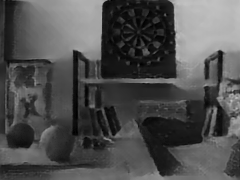} && 
    \includegraphics[width=0.23\linewidth]{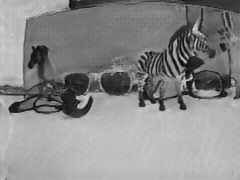} & 
    \includegraphics[width=0.23\linewidth]{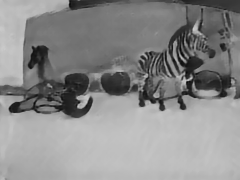} \\

    & Reconstruction & Post-Processed && Reconstruction & Post-Processed \\    
	
\end{tabular}
\caption{\textbf{Visual results of post-processing.} Here, we consider two scenes from the ECD and HQF datasets and present reconstructions of E2VID+, ET-Net, and HyperE2VID for each scene, with or without post-processing. The results demonstrate that the post-processing can satisfactorily remove or minimize most of the fine-scale artifacts, such as checkerboard patterns.}
\label{fig:post_process}
\end{figure*}